\documentclass[10pt,twocolumn,letterpaper]{article}

\usepackage{iccv}
\usepackage{times}
\usepackage{amsmath}
\usepackage{amssymb}
\usepackage{mathrsfs} 
\usepackage[standard]{ntheorem}

\usepackage{tikz}
    \definecolor{mycolor1}{HTML}{009900}
    \definecolor{mycolor2}{HTML}{000000}
    \definecolor{mycolor3}{HTML}{990000}
    \definecolor{mycolor4}{HTML}{000099}
	\usetikzlibrary{arrows}
	\usetikzlibrary{decorations.markings}
	\tikzset{
		latex-arrow/.style={
			decoration={markings,mark=at position 1 with {\arrow[scale=1.5,#1]{latex}}},
    			postaction={decorate},
    			shorten >=0.4pt
		},
  		latex-arrow/.default=black
	}
	\tikzset{
		latex-arrow-red/.style={
			decoration={markings,mark=at position 1 with {\arrow[scale=1.5,#1]{latex}}},
    			postaction={decorate},
    			shorten >=0.4pt
		},
  		latex-arrow-red/.default=mycolor1
	}
	\tikzstyle{dotnode}=[circle,fill=black,inner sep=0ex,minimum size=1.5ex]
	\tikzstyle{mynode}=[circle, draw=black, fill=white, inner sep=0pt, minimum size=2ex]
	\tikzstyle{edge}=[draw=black,latex-arrow]
	\tikzstyle{red-edge}=[draw=mycolor1,latex-arrow-red]
\usepackage{microtype}
\usepackage{pgfplots}
\usepgfplotslibrary{external}
\usepackage[ruled, vlined, linesnumbered]{algorithm2e}
    \SetAlFnt{\small}
    \SetAlCapFnt{\small}
    \SetAlCapNameFnt{\small}
\usepackage{booktabs}
\usepackage{placeins}

\usepackage[
    pdftitle={Efficient Decomposition of Image and Mesh Graphs by Lifted Multicuts},
    pdfauthor={Margret Keuper, Evgeny Levinkov, Nicolas Bonneel, Guillaume Lavoue, Thomas Brox, Bjoern Andres},
    pdfsubject={Application of Lifted Multicuts for the Decomposition of Image and Mesh Graphs},
    pdfkeywords={Lifted Multicut, Image Segmentation, Mesh Segmentation, Decomposition, Connectedness, Connectivity},
    pagebackref=true,
    breaklinks=true,
    letterpaper=true,
    colorlinks,
    bookmarks=false,
    linkcolor=mycolor1,
    urlcolor=mycolor1,
    citecolor=mycolor1
]{hyperref}

\iccvfinalcopy 

\def\iccvPaperID{2372} 

\ificcvfinal\fi
\begin{document}

\makeatletter
\providecommand*{\cupdot}{%
  \mathbin{%
    \mathpalette\@cupdot{}%
  }%
}
\newcommand*{\@cupdot}[2]{%
  \ooalign{%
    $\m@th#1\cup$\cr
    \hidewidth$\m@th#1\cdot$\hidewidth
  }%
}
\makeatother

\title{Efficient Decomposition of Image and Mesh Graphs by Lifted Multicuts}

\author{%
M.~Keuper$^1$, E.~Levinkov$^2$, N.~Bonneel$^3$, G.~Lavou\'e$^3$, T.~Brox$^1$ and B.~Andres$^{2,*}$\\[0.5ex]
\centering
\scalebox{0.8}{%
    \begin{minipage}{\textwidth}
        \centering
        $^1$\textit{Department of Computer Science, University of Freiburg}\\
        $^2$\textit{Combinatorial Image Analysis, MPI for Informatics, Saarbr\"ucken}\\
        $^3$\textit{Laboratoire d'Informatique en Image et Syst\`emes d'Information, CNRS Lyon}
    \end{minipage}
}
}

\maketitle
\renewcommand*{\thefootnote}{\fnsymbol{footnote}}
\footnotetext[1]{Correspondence: \href{mailto:andres@mpi-inf.mpg.de}{andres@mpi-inf.mpg.de}}

\newcommand{\plotNewEvaluation}[8]{
    \begin{tikzpicture}[font=\small]
    \begin{axis}[
        width=0.67\columnwidth,
        height=0.67\columnwidth,
        xlabel={VI (false cut)}, 
        ylabel={VI (false join)},
        ylabel near ticks,
        xlabel near ticks,
        xmin=0, xmax=4,
        ymin=0, ymax=4,
        mark size=0.3ex,
        legend entries={SE+multi+ucm,SE+MS+SH+ucm,#7},
        legend style={draw=none,nodes={right}} 
    ]

    \addplot[forget plot, no marks, color=black!20] {-x+1};
    \addplot[forget plot, no marks, color=black!20] {-x+2};
    \addplot[forget plot, no marks, color=black!20] {-x+3};
    \addplot[forget plot, no marks, color=black!20] {-x+4};
    \addplot[forget plot, no marks, color=black!20] {-x+5};
    \addplot[forget plot, no marks, color=black!20] {-x+6};

    \addplot[
        color=gray
    ] table [
        col sep=comma, 
        x index=6, y index=12
    ] {data-final/competing-mcg-ucm/eval_vipr.txt}; 

    \addplot[
        color=gray,
        dashed
    ] table [
        col sep=comma, 
        x index=6, y index=12
    ] {data-final/competing-se-sh-ucm/eval_vipr.txt}; 

    \addplot[
        forget plot, only marks, color=mycolor2!30,
        error bars/.cd, x dir=plus, x explicit, y dir=plus, y explicit,
    ] table [
        col sep=comma, 
        x index=6, y index=12,
        x error expr=\thisrowno{4}-\thisrowno{6}, y error expr=\thisrowno{10}-\thisrowno{12}
    ] {#1-eval_vipr.txt};

    \addplot[
        forget plot, only marks, color=mycolor2!30,
        error bars/.cd, x dir=minus, x explicit, y dir=minus, y explicit
    ] table [
        col sep=comma, 
        x index=6, y index=12,
        x error expr=-(\thisrowno{2}-\thisrowno{6}), y error expr=-(\thisrowno{8}-\thisrowno{12})
    ] {#1-eval_vipr.txt};

    \addplot[
        only marks, color=mycolor2
    ] table [
        col sep=comma, 
        x index=6, y index=12
    ] {#1-eval_vipr.txt}; 

    \ifthenelse{\equal{#3}{}}{}{%
    \addplot[
        forget plot, only marks, color=mycolor1!30,
        error bars/.cd, x dir=plus, x explicit, y dir=plus, y explicit,
    ] table [
        col sep=comma, 
        x index=6, y index=12,
        x error expr=\thisrowno{4}-\thisrowno{6}, y error expr=\thisrowno{10}-\thisrowno{12}
    ] {#3-eval_vipr.txt};

    \addplot[
        forget plot, only marks, color=mycolor1!30,
        error bars/.cd, x dir=minus, x explicit, y dir=minus, y explicit
    ] table [
        col sep=comma, 
        x index=6, y index=12,
        x error expr=-(\thisrowno{2}-\thisrowno{6}), y error expr=-(\thisrowno{8}-\thisrowno{12})
    ] {#3-eval_vipr.txt};

    \addplot[
        only marks, color=mycolor1
    ] table [
        col sep=comma, 
        x index=6, y index=12
    ] {#3-eval_vipr.txt}; 
    }

    \ifthenelse{\equal{#5}{}}{}{%
    \addplot[
        forget plot, only marks, color=mycolor3!30,
        error bars/.cd, x dir=plus, x explicit, y dir=plus, y explicit,
    ] table [
        col sep=comma, 
        x index=6, y index=12,
        x error expr=\thisrowno{4}-\thisrowno{6}, y error expr=\thisrowno{10}-\thisrowno{12}
    ] {#5-eval_vipr.txt};

    \addplot[
        forget plot, only marks, color=mycolor3!30,
        error bars/.cd, x dir=minus, x explicit, y dir=minus, y explicit
    ] table [
        col sep=comma, 
        x index=6, y index=12,
        x error expr=-(\thisrowno{2}-\thisrowno{6}), y error expr=-(\thisrowno{8}-\thisrowno{12})
    ] {#5-eval_vipr.txt};

    \addplot[
        only marks, color=mycolor3
    ] table [
        col sep=comma, 
        x index=6, y index=12
    ] {#5-eval_vipr.txt}; 
    }

    \end{axis}
    \end{tikzpicture}\hspace{0.65cm}
    \begin{tikzpicture}[font=\small]
    \begin{axis}[
        width=0.67\columnwidth,
        height=0.67\columnwidth,ylabel near ticks,
        xlabel near ticks,
        xlabel={Boundary Recall},
        ylabel={Boundary Precision},
        xmin=0, xmax=1,
        ymin=0, ymax=1,
        mark size=0.3ex,
    ]

    \addplot[forget plot, no marks, color=black!20, samples=100, domain=0.6:1] {-1/(1/x - 2/0.9)};
    \addplot[forget plot, no marks, color=black!20, samples=100, domain=0.6:1] {-1/(1/x - 2/0.8)};
    \addplot[forget plot, no marks, color=black!20, samples=100, domain=0.5:1] {-1/(1/x - 2/0.7)};
    \addplot[forget plot, no marks, color=black!20, samples=100, domain=0.4:1] {-1/(1/x - 2/0.6)};
    \addplot[forget plot, no marks, color=black!20, samples=100, domain=0.3:1] {-1/(1/x - 2/0.5)};
    \addplot[forget plot, no marks, color=black!20, samples=100, domain=0.2:1] {-1/(1/x - 2/0.4)};
    \addplot[forget plot, no marks, color=black!20, samples=100, domain=0.15:1] {-1/(1/x - 2/0.3)};
    \addplot[forget plot, no marks, color=black!20, samples=100, domain=0.1:1] {-1/(1/x - 2/0.2)};
    \addplot[forget plot, no marks, color=black!20, samples=100, domain=0.05263157894:1] {-1/(1/x - 2/0.1)};

    \addplot[
        color=gray
    ] table [
        col sep=comma, 
        x index=6, y index=12
    ] {data-final/competing-mcg-ucm/eval_pr.txt}; 

    \addplot[
        color=gray,
        dashed
    ] table [
        col sep=comma, 
        x index=6, y index=12
    ] {data-final/competing-se-sh-ucm/eval_pr.txt}; 

    \addplot[
        forget plot, only marks, color=mycolor2!30,
        error bars/.cd, x dir=plus, x explicit, y dir=plus, y explicit,
    ] table [
        col sep=comma, 
        x index=6, y index=12,
        x error expr=-(\thisrowno{2}-\thisrowno{6}), y error expr=-(\thisrowno{8}-\thisrowno{12})
    ] {#1-eval_pr.txt};

    \addplot[
        forget plot, only marks, color=mycolor2!30,
        error bars/.cd, x dir=minus, x explicit, y dir=minus, y explicit
    ] table [
        col sep=comma, 
         x index=6, y index=12,
        x error expr=-(\thisrowno{2}-\thisrowno{6}), y error expr=-(\thisrowno{8}-\thisrowno{12})
    ] {#1-eval_pr.txt};

    \addplot[
        only marks, color=mycolor2
    ] table [
        col sep=comma, 
        x index=6, y index=12
    ] {#1-eval_pr.txt}; 

    \ifthenelse{\equal{#3}{}}{}{%
    \addplot[
        forget plot, only marks, color=mycolor1!30,
        error bars/.cd, x dir=plus, x explicit, y dir=plus, y explicit,
    ] table [
        col sep=comma, 
        x index=6, y index=12,
        x error expr=-(\thisrowno{2}-\thisrowno{6}), y error expr=-(\thisrowno{8}-\thisrowno{12})
    ] {#3-eval_pr.txt};

    \addplot[
        forget plot, only marks, color=mycolor1!30,
        error bars/.cd, x dir=minus, x explicit, y dir=minus, y explicit
    ] table [
        col sep=comma, 
       x index=6, y index=12,
        x error expr=-(\thisrowno{2}-\thisrowno{6}), y error expr=-(\thisrowno{8}-\thisrowno{12})
    ] {#3-eval_pr.txt};

    \addplot[
        only marks, color=mycolor1
    ] table [
        col sep=comma, 
        x index=6, y index=12
    ] {#3-eval_pr.txt}; 
    }

    \ifthenelse{\equal{#5}{}}{}{%
    \addplot[
        forget plot, only marks, color=mycolor3!30,
        error bars/.cd, x dir=plus, x explicit, y dir=plus, y explicit,
    ] table [
        col sep=comma, 
        x index=6, y index=12,
        x error expr=-(\thisrowno{2}-\thisrowno{6}), y error expr=-(\thisrowno{8}-\thisrowno{12})
    ] {#5-eval_pr.txt};

    \addplot[
        forget plot, only marks, color=mycolor3!30,
        error bars/.cd, x dir=minus, x explicit, y dir=minus, y explicit
    ] table [
        col sep=comma, 
        x index=6, y index=12,
        x error expr=-(\thisrowno{2}-\thisrowno{6}), y error expr=-(\thisrowno{8}-\thisrowno{12})
    ] {#5-eval_pr.txt};

    \addplot[
        only marks, color=mycolor3
    ] table [
        col sep=comma, 
        x index=6, y index=12
    ] {#5-eval_pr.txt}; 
    }

    \end{axis}
    \end{tikzpicture}\hspace{0.65cm}
    \begin{tikzpicture}[font=\small]
    \begin{axis}[
        width=0.67\columnwidth,
        height=0.67\columnwidth,
        xlabel={$p^*$},
        ylabel={Runtime [s]},ylabel near ticks,
        xlabel near ticks,
        ymode=log,
        xmin=0, xmax=1,
        ymin=5e-1, ymax=1e5,
        mark size=0.3ex,
        legend entries={#8},
        legend style={draw=none,nodes={right}},
        legend pos=north west
    ]

    \addplot[
        forget plot, only marks, color=mycolor2!30,
        error bars/.cd, x dir=plus, x explicit, y dir=plus, y explicit,
    ] table [
        col sep=comma, 
        x expr=\thisrowno{0}, 
        y index=3,
        y error expr=\thisrowno{4}-\thisrowno{3}
    ] {#1-eval_runtime.txt};

    \addplot[
        forget plot, only marks, color=mycolor2!30,
        error bars/.cd, x dir=minus, x explicit, y dir=minus, y explicit
    ] table [
        col sep=comma, 
        x expr=\thisrowno{0}, 
        y index=3,
        y error expr=\thisrowno{3}-\thisrowno{2}
    ] {#1-eval_runtime.txt};

    \addplot[
       only marks, color=mycolor2
    ] table [
        col sep=comma, 
        x expr=\thisrowno{0}, 
        y index=3
    ] {#1-eval_runtime.txt}; 

    \ifthenelse{\equal{#3}{}}{}{%
    \addplot[
        forget plot, only marks, color=mycolor1!30,
        error bars/.cd, x dir=plus, x explicit, y dir=plus, y explicit,
    ] table [
        col sep=comma, 
        x expr=\thisrowno{0}, 
        y index=3,
        y error expr=\thisrowno{4}-\thisrowno{3}
    ] {#3-eval_runtime.txt};

    \addplot[
        forget plot, only marks, color=mycolor1!30,
        error bars/.cd, x dir=minus, x explicit, y dir=minus, y explicit
    ] table [
        col sep=comma, 
        x expr=\thisrowno{0}, 
        y index=3,
        y error expr=\thisrowno{3}-\thisrowno{2}
    ] {#3-eval_runtime.txt};

    \addplot[
       only marks, color=mycolor1
    ] table [
        col sep=comma, 
        x expr=\thisrowno{0}, 
        y index=3
    ] {#3-eval_runtime.txt}; 
    }

    \ifthenelse{\equal{#5}{}}{}{%
    \addplot[
        forget plot, only marks, color=mycolor3!30,
        error bars/.cd, x dir=plus, x explicit, y dir=plus, y explicit,
    ] table [
        col sep=comma, 
        x expr=\thisrowno{0}, 
        y index=3,
        y error expr=\thisrowno{4}-\thisrowno{3}
    ] {#5-eval_runtime.txt};

    \addplot[
        forget plot, only marks, color=mycolor3!30,
        error bars/.cd, x dir=minus, x explicit, y dir=minus, y explicit
    ] table [
        col sep=comma, 
        x expr=\thisrowno{0}, 
        y index=3,
        y error expr=\thisrowno{3}-\thisrowno{2}
    ] {#5-eval_runtime.txt};

    \addplot[
       only marks, color=mycolor3
    ] table [
        col sep=comma, 
        x expr=\thisrowno{0}, 
        y index=3
    ] {#5-eval_runtime.txt}; 
    }

    \end{axis}
    \end{tikzpicture}
}

\begin{abstract}
Formulations of the Image Decomposition Problem
\cite{arbelaez-2011}
as a Multicut Problem (MP) w.r.t.~a superpixel graph have received considerable attention.
In contrast, instances of the MP w.r.t.~a pixel grid graph have received little attention,
firstly, because the MP is NP-hard and instances w.r.t.~a pixel grid graph are hard to solve in practice, and,
secondly, due to the lack of long-range terms in the objective function of the MP.
We propose a generalization of the MP with long-range terms (LMP).
We design and implement two efficient algorithms (primal feasible heuristics) for the MP and LMP which allow us to study instances of both problems w.r.t.~the pixel grid graphs of the images in the BSDS-500 benchmark
\cite{arbelaez-2011}.
The decompositions we obtain do not differ significantly from the state of the art, suggesting that the LMP is a competitive formulation of the Image Decomposition Problem.
To demonstrate the generality of the LMP, 
we apply it also to the Mesh Decomposition Problem posed by the Princeton benchmark
\cite{Chen2009},
obtaining state-of-the-art decompositions.
\end{abstract}

\section{Introduction}

Formulations of the Image Decomposition Problem
\cite{arbelaez-2011}
as a Minimum Cost Multicut Problem (MP)
\cite{chopra-1993,deza-1997}
have received considerable attention 
\cite{alush-2012,andres-2011,andres-2012,andres-2013,bagon-2011,beier-2015,beier-2014,kappes-2011,kappes-2013-arxiv,kappes-2015-ssvm,kim-2011,kim-2014,nowozin-2009,yarkony-2012,yarkony-2015}.
Advantages of this formulation are in order:
Firstly, the feasible solutions of the MP relate one-to-one to the decompositions of a graph.
In particular, the number of components is not fixed in advance but is determined by the solution.
Secondly, the MP, unlike balanced cut problems 
\cite{shi-2000},
does not favor one decomposition over another by definition.
Thirdly, multicut algorithms are easy to use;
they take as input a graph, 
\eg the pixel grid graph of an image, 
and, for every edge, a real-valued cost (reward) of the incident nodes being in distinct components,
\eg $\log \frac{1-p_e}{p_e} + \log \frac{1-p^*}{p^*}$, for an estimated probability $p_e$ of boundary 
\cite{arbelaez-2011}
at the edge $e$, and a prior probability $p^* \in (0, 1)$ of cuts.
The output is a 01-labeling of the edges 
that well-defines a decomposition of the graph 
by $0$ indicating ``join'' and $1$ indicating ``cut''.

One disadvantage is the NP-hardness of the MP 
\cite{bansal-2004,demaine-2006}.
Despite significant progress in the design of efficient heuristics
\cite{bagon-2011,beier-2015,beier-2014,kernighan-1970},
instances of the MP for image segmentation have so far only been solved w.r.t.~superpixel adjacency graphs and not w.r.t.~pixel grid graphs, with the sole and notable exception of
\cite{bagon-2011}.
A second disadvantage results from the fact that a multicut makes explicit only for edges whether the incident nodes are in distinct components.
It does not make explicit for pairs of nodes that are not neighbors whether these are in distinct components.
Hence, the linear objective function of the MP w.r.t.~a pixel grid graph cannot assign a cost specifically to all decompositions for which a pair of pixels that are not neighbors are in distinct components.
This limitation, noted \eg in 
\cite{andres-2013},
hampers applications as it is often hard to estimate, for an image and a pair of neighboring pixels, whether the image is to be cut precisely between these pixels (only these estimates are used in the MP), and as it is sometimes easy to estimate for pixels at larger distance whether these are in distinct components (these estimates are not used in the MP).

An optimization problem whose feasible solutions relate one-to-one to the decompositions of a graph 
and whose objective function can assign, for any pair of nodes, a cost to all decompositions for which these nodes are in distinct components, although desirable, has not been proposed before.

\textbf{Contribution.}
We propose the Minimum Cost Lifted Multicut Problem (LMP),
a generalization of the MP whose feasible solutions relate one-to-one to the decompositions of a graph and whose objective function can assign, for any pair of nodes, a real-valued cost (reward) to all decompositions for which these nodes are in distinct components.
We design and implement two efficient algorighms for both the MP and the LMP
and evaluate both problem formulations in conjunction with both algorithms 
for the Image Decomposition Problem in terms of the BSDS-500 benchmark
\cite{arbelaez-2011}
and for the Mesh Decomposition Problem in terms of the Princeton Mesh Segmentation benchmark
\cite{Chen2009}.
\vfill

\section{Related Work}

The MP is known as Correlation Clustering in machine learning 
and theoretical computer science
\cite{bansal-2004,demaine-2006}.
For complete graphs, which are of special interest in machine learning, the well-known MP and the proposed LMP coincide.

A generalization of the MP by a higher-order objective function, called the Higher-Order Multicut Problem (HMP), was proposed in
\cite{kim-2011}
and is studied in detail in
\cite{kappes-2013-arxiv,kim-2014}.
In principle, the HMP subsumes all optimization problems whose feasible solutions coincide with the multicuts of a graph, including the LMP we propose.
In fact, the HMP is strictly more general than the LMP; its objective function can assign an objective value to all decompositions for which any set of edges is cut, unlike the objective function of the LMP which is limited to single edges.
However, the instances of the HMP that are equivalent to the instances of the LMP we propose have an objective function whose order is equal to the number of edges in the graph
and are hence impractical.
Thus, the HMP and LMP are complementary in practice.

Efficient algorithms (primal feasible heuristics) for the MP are proposed and analyzed in \cite{bagon-2011,beier-2014,beier-2015,kernighan-1970}.
The algorithms we design and implement are compared here to the state of the art
\cite{beier-2014}.
Our implementation of (an extension of) the Kernighan-Lin Algorithm (KL)
\cite{kernighan-1970}
is compared here, in addition, to the implementation of KL in 
\cite{andres-2012-opengm,kappes-2015}.

Toward image decomposition
\cite{arbelaez-2011},
the state of the art in boundary detection is
\cite{bertrasius-2015, fowlkes-2015},
followed closely by
\cite{DollarICCV13edges, crisp_boundaries}.
Our experiments are based on
\cite{DollarICCV13edges}
which is publicly available and outperformed marginally by
\cite{bertrasius-2015, fowlkes-2015}.
The state of the art in image decomposition is
\cite{APBMM2014},
followed closely by
\cite{arbelaez-2011, crisp_boundaries}.
Our results are compared quantitatively to
\cite{APBMM2014}.

Toward mesh decomposition
\cite{Theologou2015},
the state of the art is 
\cite{Kalogerakis2010,Benhabiles2011,Xie2014}, 
followed closely by
\cite{Zhanga,Kin-ChungAu2011}.
Our experiments are based on
\cite{Kalogerakis2010,Zhanga,Kin-ChungAu2011}.
In prior work, methods based on learning mostly rely on a unary term which requires components to be labeled semantically
\cite{Kalogerakis2010,Xie2014}. 
One method based on edge probabilities was introduced previously \cite{Benhabiles2011}. 
It applies a complex post-process (contour thinning and completion, snake movement) to obtain a decomposition. 
We show the first mesh decompositions based on multicuts.

\section{Problem Formulation}
\label{section:problem-formulation}

\subsection{Minimum Cost Lifted Multicut Problem}
\label{section:model}
We now define an optimization problem,
the Minimum Cost Lifted Multicut Problem,
whose feasible solutions relate one-to-one to the decompositions of a graph
and whose objective function can assign, for any pair of nodes, a cost to all decompositions for which these nodes are in distinct components.
Here, a component of a graph is any non-empty subgraph that is node-induced and connected.
A decomposition of a graph is any partition $\Pi$ of the node set such that, for every $V' \in \Pi$, the subgraph induced by $V'$ is connected (and hence a component of the graph).
An instance of the problem is defined w.r.t.:
\begin{itemize}
\itemsep0ex
\parskip1ex
\item A simple, undirected graph $G = (V, E)$, \eg, the pixel grid graph of an image or the triangle adjacency graph of a mesh.

\item Additional edges $F \subseteq \binom{V}{2} \setminus E$ connecting nodes that are not neighbors in $G$.
In practice, we choose $F$ so as to connect any two nodes $v,w \in V$ whose distance $d_{vw}$ in the graph holds $1 < d_{vw} \leq d^*$ for a maximum distance $d^* \in \mathbb{R}_0^+$,
fixed for the experiments in 
Sec.~\ref{section:experiments}.

\item For every edge $vw \in E \cup F$, a cost $c_{vw} \in \mathbb{R}$
assigned to all feasible solutions for which $v$ and $w$ are in distinct components.
The estimation of $c_{vw}$ from image and mesh data is discussed in 
Sections~\ref{section:probability} and \ref{section:experiments}.
\end{itemize}

With respect to the above, we define a feasible set $Y_{EF} \subseteq \{0,1\}^{E \cup F}$ whose elements $y \in Y_{EF}$
are 01-labelings of all edges $E \cup F$.
The feasible set is defined such that two conditions hold:
Firstly, the feasible solutions $y \in Y_{EF}$ relate one-to-one to the decompositions of the graph $G$. 
Secondly, for every edge $vw \in E \cup F$, $y_{vw} = 1$ if and only if $v$ and $w$ are in distinct components of $G$.
This is expressed rigorously by two classes of constraints:
The linear inequalities \eqref{eq:lmc-cycle} below constrain $y$ such that
$\{e \in E\ |\ y_e = 1\}$ is a multicut of the graph $G$
\cite{chopra-1993}.
For any decomposition of a graph, the multicut related to the decomposition is the subset of those edges that straddle distinct components.
In addition, the linear inequalities \eqref{eq:lmc-path} and \eqref{eq:lmc-cut} constrain $y$ such that, for any $vw \in F$, $y_{vw} = 0$ if and only if there exists a path in $G$ from $v$ to $w$, along which all edges are labeled $0$.
%
\begin{definition}
\label{definition:problem}
For any simple, undirected graph $G = (V, E)$,
any $F \subseteq \binom{V}{2} \setminus E$
and any  $c: E \cup F \to \mathbb{R}$,
the 01 linear program written below is called an instance of the
\emph{Minimum Cost Lifted Multicut Problem (LMP)}
w.r.t.~$G$, $F$ and $c$.
\begin{align}
\min_{y \in Y_{EF}} 
    & \sum_{e \in E \cup F} c_e y_e
    \label{eq:problem}
\end{align}
with $Y_{EF} \subseteq \{0,1\}^{E \cup F}$ the set of all $y \in \{0,1\}^{E \cup F}$ with
\begin{align}
& \hspace{-1ex} \forall C \in \textnormal{cycles}(G)\ \forall e \in C:\ 
    y_e \leq \hspace{-2ex} \sum_{e' \in C \setminus \{e\}} \hspace{-2ex} y_{e'} &
    \label{eq:lmc-cycle}\\
& \hspace{-1ex} \forall vw \in F\ \forall P \in vw\textnormal{-paths}(G):\ 
    y_{vw} \leq \sum_{e \in P} y_e &
    \label{eq:lmc-path}\\
& \hspace{-1ex} \forall vw \in F\ \forall C \in vw\textnormal{-cuts}(G):
    1 - y_{vw} \leq \sum_{e \in C} (1 - y_e) 
    \label{eq:lmc-cut}
\end{align}
\end{definition}

\subsection{Properties}

We now discuss properties of the LMP 
(Def.~\ref{definition:problem}):

For $F = \emptyset$, the LMP specializes to the MP
\cite{chopra-1993,deza-1997}.
Its feasible set $Y_{E\emptyset}$ consists of the characteristic functions of all multicuts of $G$
(which relate one-to-one to the decompositions of $G$).
Its linear objective function can be chosen so as to assign, for any edge $vw \in E$, a cost $c_{vw} \in \mathbb{R}$  to all decompositions of $G$ for which the nodes $v$ and $w$ are in distinct components.
It cannot be chosen so as to assign, for distinct nodes $v$ and $w$ that are not neighbors in $G$, a cost precisely to all decompositions of $G$ for which $v$ and $w$ are in distinct components.

For $F \neq \emptyset$, the LMP is not a MP.
Its feasible solutions still relate one-to-one to the decompositions of the graph $G$
(because $\varphi: Y_{EF} \to Y_{E\emptyset}: y \mapsto y_E$ is a bijection).
Its objective function can be chosen so as to assign, for any $vw \in E \cup F$, a cost to all decompositions for which the nodes $v$ and $w$ are in distinct components.
Thus, the LMP generalizes the MP.
The feasible solutions $y \in Y_{EF}$ are called \emph{lifted multicuts} from $(V, E)$ to $(V, E \cup F)$ and are studied in
\cite{andres-2015}.

For some instances of the LMP, 
notably if $c_F < 0$
\cite{andres-2013}, 
its solutions can be identified with the solutions of the instance of the MP w.r.t.~the larger graph $G' := (V, E \cup F)$ and $c$.
For the general LMP, this is not true.
The feasible solutions of the MP with respect to $G'$ and $c$ do not relate one-to-one to the decompositions of $G$,
unlike the feasible solutions of the LMP which are the characteristic functions of \emph{some} multicuts of $G'$, namely those that are \emph{lifted} from $G$
\cite{andres-2015}.

A cutting plane algorithm for the LMP, based on the canonical LP-relaxation of the ILP in  
Def.~\ref{definition:problem},
is impractical for the instances we consider in 
Sec.~\ref{section:experiments}:
Although the inequalities \eqref{eq:lmc-cycle}--\eqref{eq:lmc-cut} can be separated efficiently,
the number of to-be-separated inequalities \eqref{eq:lmc-cut} is prohibitive,
and the facet-defining subset of \eqref{eq:lmc-cut} is unknown
\cite{andres-2015}.
Thus, we propose in 
Sec.~\ref{section:algorithms}
two primal feasible heuristics for the LMP.

\subsection{Probabilistic Model}
\label{section:probability}

\begin{figure}
\centering
\begin{tikzpicture}[font=\small]
    \node[style=mynode, label=left:$\mathcal{X}_e$] (x) at (1.5, 2) {};
    \node[style=mynode, label=left:$\mathcal{Y}_e$] (y) at (1.5, 1) {};
    \node[style=mynode, label=left:$\mathscr{Y}$] (z) at (1.5, -0.5) {};
    \node[style=mynode, draw=mycolor1, label=right:$\color{mycolor1}\mathcal{Y}_f$] (y2) at (3, 1) {};
    \draw (0.35, 0.2) rectangle (2, 2.5);
    \draw[mycolor1] (2.3, 0.2) rectangle (4.1, 1.6);
    \node at (0.95, 0.5) {$e \in E$};
    \path[style=edge] (x)--(y) {};
    \path[style=edge] (y)--(z) {};
    \path[style=red-edge] (y2)--(z) {};
    \path[style=red-edge] (x)--(y2) {};
    \node at (3.5, 0.5) {\color{mycolor1}$f \in F$};
\end{tikzpicture}\\[1ex]
\caption{Depicted in black is a Bayesian Network defining a set of probability measures on multicuts
\cite{andres-2011}.
Depicted in green is our extension defining a set of probability measures on lifted multicuts.}
\label{figure:bayesian-network}
\end{figure}
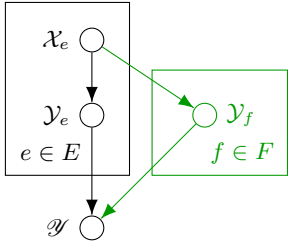

We now define a familiy of probability measures on lifted multicuts
for which the maximally probable lifted multicuts are the solutions the LMP
(Def.~\ref{definition:problem}).
This relates the coefficients $c$ of the LMP to image and mesh data.

\textbf{Probability measures on multicuts.}
Andres et al.~\cite{andres-2011} 
define, with respect to a graph $G = (V, E)$ and with respect to the Bayesian Network depicted in
Fig.~\ref{figure:bayesian-network} (in black),
a measure of the conditional probability of a $y \in \{0,1\}^E$, 
given the feasible set $Y_{E \emptyset}$ of the characteristic functions of all multicuts of $G$ and
given, for every edge $e \in E$, a vector $x_e \in \mathbb{R}^n$ of $n \in \mathbb{N}$ edge features.
Specifically,
\begin{align}
p_{\mathcal{Y} | \mathcal{X}, \mathscr{Y}} 
\ & \propto \ 
p_{\mathscr{Y} | \mathcal{Y}} \cdot \prod_{e \in E} p_{\mathcal{Y}_e | \mathcal{X}_e}
\label{eq:full-measure}\\
\textnormal{with} \qquad
p_{\mathscr{Y} | \mathcal{Y}}(Y_{E \emptyset}, y)
\ & \propto \ 
\begin{cases}
1 & \textnormal{if}\ y \in Y_{E \emptyset}\\
0 & \textnormal{otherwise}
\end{cases}
\enspace .
\label{eq:renormalization}
\end{align}

They show that $y$ maximizes $p_{\mathcal{Y} | \mathcal{X}, \mathscr{Y}}$ if and only if it is a solution of the instance of the MP with respect to $G$ and $c \in \mathbb{R}^E$ such that
\begin{align}
\forall e \in E: \quad
c_e = \log \frac{p_{\mathcal{Y}_e | \mathcal{X}_e}(0, x_e)}{p_{\mathcal{Y}_e | \mathcal{X}_e}(1, x_e)}
\enspace .
\label{eq:prob-map}
\end{align}

\textbf{Probability measures on lifted multicuts.}
We extend the Bayesian Network of 
Andres et al.~\cite{andres-2011}
in order to incorporate estimated probabilities not only for edges but also for pairs of nodes that are not neighbors.

The extension is depicted in 
Fig.~\ref{figure:bayesian-network}
(in green).
It contains one additional random variable $\mathcal{Y}_f$ for every $f \in F$.
The conditional probability measures $p_{\mathcal{Y} | \mathcal{X}, \mathscr{Y}}$ 
consistent with the extended Bayesian Network have the form 
\begin{align}
p_{\mathcal{Y} | \mathcal{X}, \mathscr{Y}}
    \propto 
        p_{\mathscr{Y} | \mathcal{Y}} 
        \cdot \prod_{e \in E} p_{\mathcal{Y}_e | \mathcal{X}_e}
        \cdot \prod_{f \in F} p_{\mathcal{Y}_f | \mathcal{X}_E}
\enspace .
\end{align}

A realization of all random variables $\mathcal{Y}$ is a 01-labeling $y \in \{0,1\}^{E \cup F}$ of all edges $vw \in E \cup F$.
In order to constrain it to the characteristic functions of lifted multicuts, 
we consider
\eqref{eq:renormalization}
with $Y_{EF}$ instead of $Y_{E\emptyset}$.

\textbf{Probabilistic Geodesic Lifting.}
Estimating, for edges $vw = e \in E$, the probability $p_{\mathcal{Y}_e | \mathcal{X}_e}$ of the nodes $v$ and $w$ being in distinct components, given features $x_e$ defined by image and mesh data, is the classical problem of boundary estimation
\cite{arbelaez-2011}.
In our experiments described in 
Sec.~\ref{section:experiments},
we build on recent work 
\cite{DollarICCV13edges,Kalogerakis2010,Kin-ChungAu2011,Zhanga}
in this field.

Estimating, for pairs $vw = f \in F$ of nodes $v$ and $w$ that are not neighbors, 
the probability $p_{\mathcal{Y}_f | \mathcal{X}_E}$ of $v$ and $w$ being in distinct components
is a much harder problem:
As these nodes could be connected by any path in $G$, this probability depends on the features $x_E$ of all edges.
In our experiments,
we define, for all $vw = f \in F$:
\begin{align}
p_{\mathcal{Y}_f | \mathcal{X}_E}(0, x_E) 
:= \max_{P \in vw\textnormal{-paths}(G)} \prod_{e \in P} p_{\mathcal{Y}_e | \mathcal{X}_e}(0, x_e) 
\enspace .
\end{align}

On the one hand, this under-estimates the probability as only one path is considered.
On the other hand, it is the largerst such under-estimate as a maximally probable such path is considerd.
Note also that $-\log p_{\mathcal{Y}_f | \mathcal{X}_E}(0, x_E)$ can be computed efficiently using, \eg, Dijkstra's algorithm.

\section{Efficient Algorithms}
\label{section:algorithms}

We now introduce two efficient algorithms (primal feasible heuristics) 
which are applicable 
to the LMP (Def.~\ref{definition:problem})
and the MP (the special case of the LMP for $F = \emptyset$).

Alg.~\ref{alg:AGG}
is an adaptation of greedy agglomeration, more specifically, greedy additive edge contraction.
It takes as input an instance of the LMP
defined by $G = (V,E)$, $F$ and $c$
(Def.~\ref{definition:problem})
and constructs as output a decomposition of the graph $G$.
Alg.~\ref{alg:KL_outer}
is an extension of the Kernighan-Lin Algorithm 
\cite{kernighan-1970}.
It takes as input an instance of the LMP and an initial decomposition of $G$ and constructs as output a decomposition of $G$ whose lifted multicut has an objective value lower than or equal to that of the initial decomposition. 
Both algorithms maintain a decomposition of $G$,
represented by graph $\mathcal{G} = (\mathcal{V}, \mathcal{E})$
whose nodes $a \in \mathcal{V}$ are components of $G$
and whose edges $ab \in \mathcal{E}$ connect any components $a$ and $b$ of $G$ which are neighbors in $G$.
Objective values are computed w.r.t.~the larger graph $G' = (V, E \cup F)$ and $c$.

\subsection{Greedy Additive Edge Contraction}

\begin{algorithm}[t]
\caption{Greedy Additive Edge Contraction (GAEC)}
\label{alg:AGG}
\DontPrintSemicolon
\While{$\mathcal{E} \neq \emptyset$}
{
$ab := \underset{a'b' \in \mathcal{E}}{\mathrm{argmax}}\ \chi_{a'b'}$\;
\If{$\chi_{ab} < 0$}
{
  {\bf break}\;
}
\textbf{contract} $ab$ in $\mathcal{G}$ and $\mathcal{G}'$ \;
\ForEach{$ab \neq ab' \in \mathcal{E}'$}
{
    $\chi_{ab'} := \chi_{ab'} + \chi_{bb'}$\;
}
}
\end{algorithm}

\textbf{Overview.}
Alg.~\ref{alg:AGG} starts from the decomposition into single nodes.
In every iteration, a pair of neighboring components is joined for which the join decreases the objective value maximally.
If no join strictly decreases the objective value, the algorithm terminates.

\textbf{Implementation.}
Our implementation
\cite{our-code}
uses \emph{ordered adjacency lists} for the graph $\mathcal{G}$
and for a graph $\mathcal{G}' = (\mathcal{V}, \mathcal{E}')$ whose edges $ab \in \mathcal{E}'$ connect any components $a$ and $b$ of $G$ for which there is an edge $vw \in E \cup F$ with $v \in a$ and $w \in b$.
It uses a \emph{disjoint set data structure} for the partition of $V$
and a \emph{priority queue} for an ordered sequence of costs $\chi: \mathcal{E} \to \mathbb{R}$ of feasible joins.
Its worst-case time complexity $O(|V|^2 \log |V|)$ is due to 
a sequence of at most $|V|$ contractions,
in each of which at most $\deg \mathcal{G}' \leq |V|$ edges are removed,
each in time $O(\log \deg \mathcal{G}') \in O(\log |V|)$.

\subsection{Kernighan-Lin Algorithm with Joins}

\begin{algorithm}[t]
\caption{Kernighan-Lin Algorithm with Joins (KLj)}
\label{alg:KL_outer}
\DontPrintSemicolon
\SetKwFor{Forever}{repeat}{}{}
\Repeat{no changes}{
    \ForEach{$ab \in \mathcal{E}$}{
        \If{\textnormal{has\_changed}($a$) {\bf or} \textnormal{has\_changed}($b$)}{
        update\_bipartition($\mathcal{G}, a, b$)\;
        }
    }
    \ForEach{$a \in \mathcal{V}$}{
        \If{\textnormal{has\_changed}(a)}{
            \Repeat{no changes}{
              update\_bipartition($\mathcal{G}, a, \emptyset$)\;
            }
        }   
    }
}
\end{algorithm}

\textbf{Overview.}
Alg.~\ref{alg:KL_outer}
starts from an initial decomposition provided as input.
In each iteration, an attempt is made to improve the current decomposition by one of the following transformations:
1.~moving nodes between two neighboring components,
2.~moving nodes from one component to an additional, newly introduced component,
3.~joining two neighboring components.
The main operation
``update\_bipartition''
is described below.
It takes as input the current decomposition and a pair $ab \in \mathcal{E}$ of neighboring components of $G$
and assesses Transformations 1 and 3 for this pair.
Transformations 2 are assessed by executing ``update\_bipartition'' for each component and $\emptyset$.

The operation ``update\_bipartition''
constructs a sequence of elementary transformations of the components $a$ and $b$ and a $k \in \mathbb{N}_0$ such that the first $k$ elementary transformations in the sequence, carried out in order, descrease the objective value maximally.
Each elementary transformation consists in either moving a node currently in the component $a$ which currently has a neighbor in the component $b$ from $a$ to $b$,
or in moving a node currently in the component $b$ which currently has a neighbor in the component $a$ from $b$ to $a$.
The sequence of elementary transformations is constructed greedily, always choosing one elementary transformation that decreases the objective function maximally.
If either the first $k$ elementary transformations together or a complete join of the components $a$ and $b$ strictly decreases the objective value, an optimal among these operations is carried out.

\textbf{Implementation.}
Our implementation
\cite{our-code}
of Alg.~\ref{alg:KL_outer}
tags components that are updated
in order to avoid 
that a pair of components that is fixed under ``update\_bipartition'' is processed more than once.
In the operation ``update\_bipartition'',
we maintain the set $\Omega \subseteq E$ of edges of $G$ that straddle the components $a$ and $b$.
A substantial complication in the case of an LMP ($F \neq \emptyset$) arises from the fact that moving a node $v \in V$ from a component $a \subseteq V$ to a neighboring component $b \subseteq V$ might leave the set $a \setminus \{v\}$ disconnected.
Keeping track of these cut-vertices by the Hopcroft-Tarjan Algorithm
\cite{hopcroft-1973}
turned out to be impractical due to excessive absolute runtime.
Our implementation allows for elementary transformations that leave components disconnected;
we even compute the difference to the objective value incorrectly in such a case while constructing the sequence of elementary transformations.
However, the first $k$ elementary transformations are carried out only if the correct difference to the objective value (computed after the construction of the entire sequence) is optimal.
Our implementation of ``update\_bipartition'' has the worst-case time complexity 
$O(|a \cup b|(|\Omega| + \deg G'))$.
The number of outer iterations of 
Alg.~\ref{alg:KL_outer} 
is not bounded here by a polynomial but is typically small
(less than 20 for all experiments in 
Sec.~\ref{section:experiments}).

\section{Experiments}
\label{section:experiments}

\subsection{Image Decomposition}

We now apply both formulations of the graph decomposition problem,
the Minimum Cost Multicut Problem (MP) 
\cite{chopra-1993}
and the Minimum Cost Lifted Multicut Problem (LMP) defined in
Sec.~\ref{section:problem-formulation},
in conjunction with both algorithms defined in 
Sec.~\ref{section:algorithms},
GAEC and KLj,
to the Image Decomposition Problem posed by the BSDS-500 benchmark
\cite{arbelaez-2011}.

\begin{figure*}
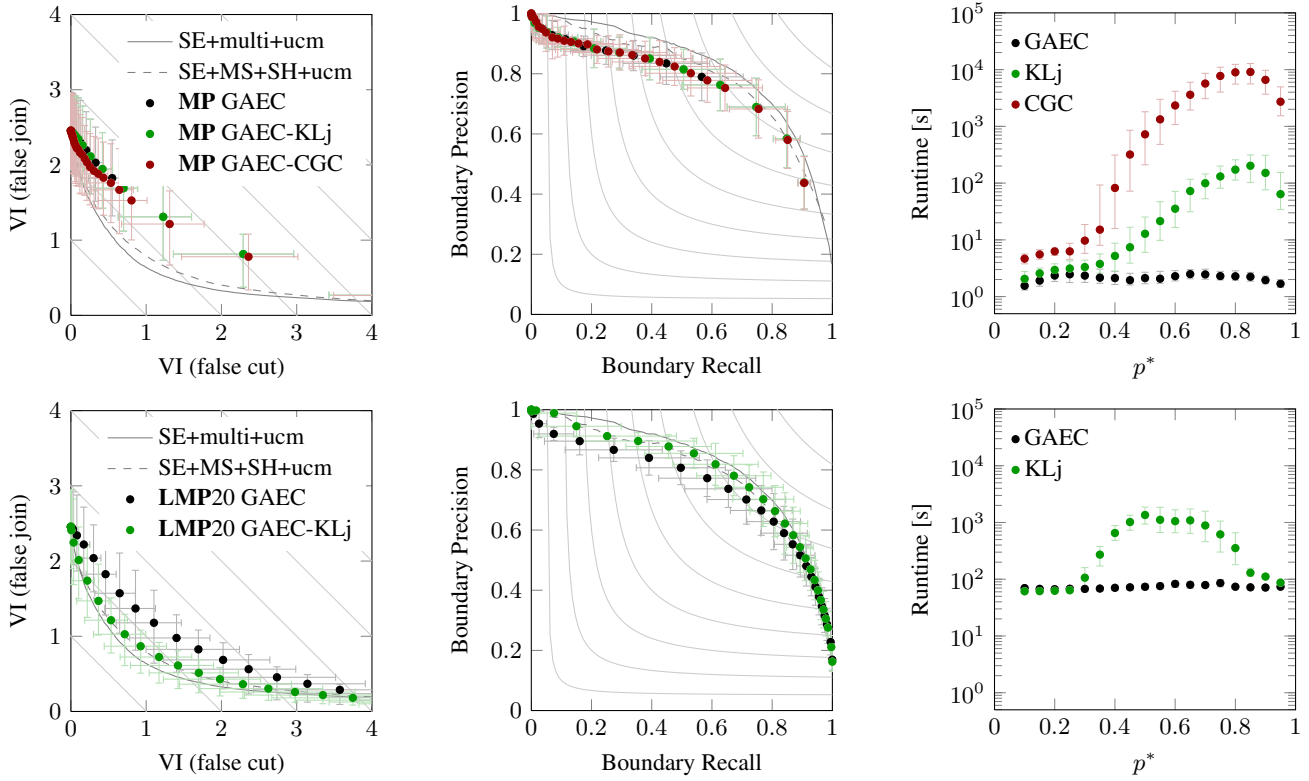

\centering
\plotNewEvaluation
    {data-final/postICCVtestwta_AGG_multicut_keep_old_weights}{}
    {data-final/postICCVtestwta_KL_AGG_multicut}{}
    {data-final/postICCVtestwta_CGC_keep_old_weights}{}
    {\textbf{MP} GAEC,\textbf{MP} GAEC-KLj,\textbf{MP} GAEC-CGC}{GAEC,KLj,CGC}
\plotNewEvaluation
    {data-final/postICCVtestwta20_AGG_multicut_keep_old_weights}{}
    {data-final/postICCVtestwta20_KL_AGG_multicut_keep_old_weights}{}
    {}{}
    {\textbf{LMP}20 GAEC,\textbf{LMP}20 GAEC-KLj}{GAEC,KLj}
\vspace{-0.4cm}
\caption{Depicted above is an assessment of 
the Multicut Problem (MP) and 
the Lifted Multicut Problem with $d^*=20$ (LMP20)
in conjunction with 
Alg.~\ref{alg:AGG} (GAEC) and 
Alg.~\ref{alg:KL_outer} (KLj),
in an application to the image decomposition problem posed by the BSDS-500 benchmark
\cite{arbelaez-2011}.
Every point in the figures above shows, for one problem and algorithm, the average over all test images in the benchmark.
Depicted are, 
on the \textbf{left}, the variation of information (VI), split additively into a distance due to false cuts and a distance due to false joins, 
in the \textbf{middle}, the accuracy of boundary detection, split into recall and precision and,
on the \textbf{right}, the absolute runtime.
The state of the art 
SE+multi+ucm \cite{APBMM2014}
and 
SE+MS+SH\cite{DollarICCV13edges}+ucm
are depicted as solid/dashed gray lines.
Error bars depict the 0.25 and 0.75-quantile.}
\label{figure:results}
\end{figure*}

For every test image, we define instances of the MP and the LMP
as descibed in
Sec.~\ref{section:problem-formulation}.
For each of these, we compute a feasible solution,
firstly, by greedy additive edge contraction (GAEC, Alg.~\ref{alg:AGG}) and,
secondly, by applying the extended Kernighan-Lin Algorithm (KLj, Alg.~\ref{alg:KL_outer}) to the output of GAEC.
All decompositions obtained in this way are compared to the man-made decompositions in the BSDS-500 benchmark 
in terms of boundary precision and recall (BPR) 
\cite{arbelaez-2011}
and variation of information (VI)
\cite{meila-2007}.
The VI is split into a distance due to false joins, plus a distances due to false cuts, as in
\cite{keuper-2015b}.
Statistics for the entire BSDS-500 test set are shown in
Tab.~\ref{table:results} and Fig.~\ref{figure:results}
and are discussed below, after a specification of the experimental setup.

\textbf{Setup.}
For every image, instances of the MP are defined w.r.t.:
1.~the pixel grid graph of the image,
2.~for every edge in this graph, \ie, for every pair of pixels that are 4-neighbors, 
the probability estimated in
\cite{DollarICCV13edges} 
of these pixels being in distinct components,
3.~a prior probability $p^*$ of neighboring pixels being in distinct components.
We vary $p^* \in \{0.05, 0.10, \ldots, 0.95\}$,
constructing one instance of the MP for every image and every $p^*$.
For each of these instance of the MP, three instances of the LMP are defined by Probabilistic Geodesic Lifting
(Sec.~\ref{section:probability}),
one for each $d^* \in \{5, 10, 20\}$.
Each experiment described in this section is conducted using one Intel Xeon CPU E5-2680 operating at $2.70$~GHz (no parallelization).

\textbf{Results.}
It can be seen form
Fig.~\ref{figure:results}
that the algorithms GAEC and KLj defined in 
Sec.~\ref{section:algorithms}
terminate in a time in the order of $10^3$ seconds for every instance of the MP and LMP we define,
more than an order of magnitude faster than the state of the art
\cite{beier-2014}.
Note that we do use the most efficient algorithm of 
\cite{beier-2014}
which exploits the planarity of the pixel grid graph.
A more detailed comparison of KLj with CGC 
\cite{beier-2014}
and the implementation of KL in
\cite{andres-2012-opengm,kappes-2015}
in terms of objective value and runtime is depicted in
Fig.~\ref{figure:algorithm-comparison}.
It can be seen from this figure that our implementation of KLj is faster than the implementation in
\cite{andres-2012-opengm,kappes-2015}
also by more than an order of magnitude.
This improvement in runtime facilitates our study of the MP and LMP with respect to the pixel grid graphs of the images in the BSDS-500 benchmark.

It can also be seen from 
Fig.~\ref{figure:results}
that feasible solutions of the MP found by GAEC are not improved significantly by either of the local search heuristics KLj or CGC 
\cite{beier-2014}.
Compared to the man-made decompositions in the benchmark in terms of BPR and VI,
feasbile solutions of the MP found by GAEC, improved by either CGC or KLj,
are significantly worse than the state of the art
\cite{APBMM2014}
for this benchmark 
(Tab.~\ref{table:results}).

\begin{figure*}
\centering
\small
\begin{tabular}{@{}c@{}c@{}c@{}c@{}c@{}}
\includegraphics[width=0.195\textwidth]{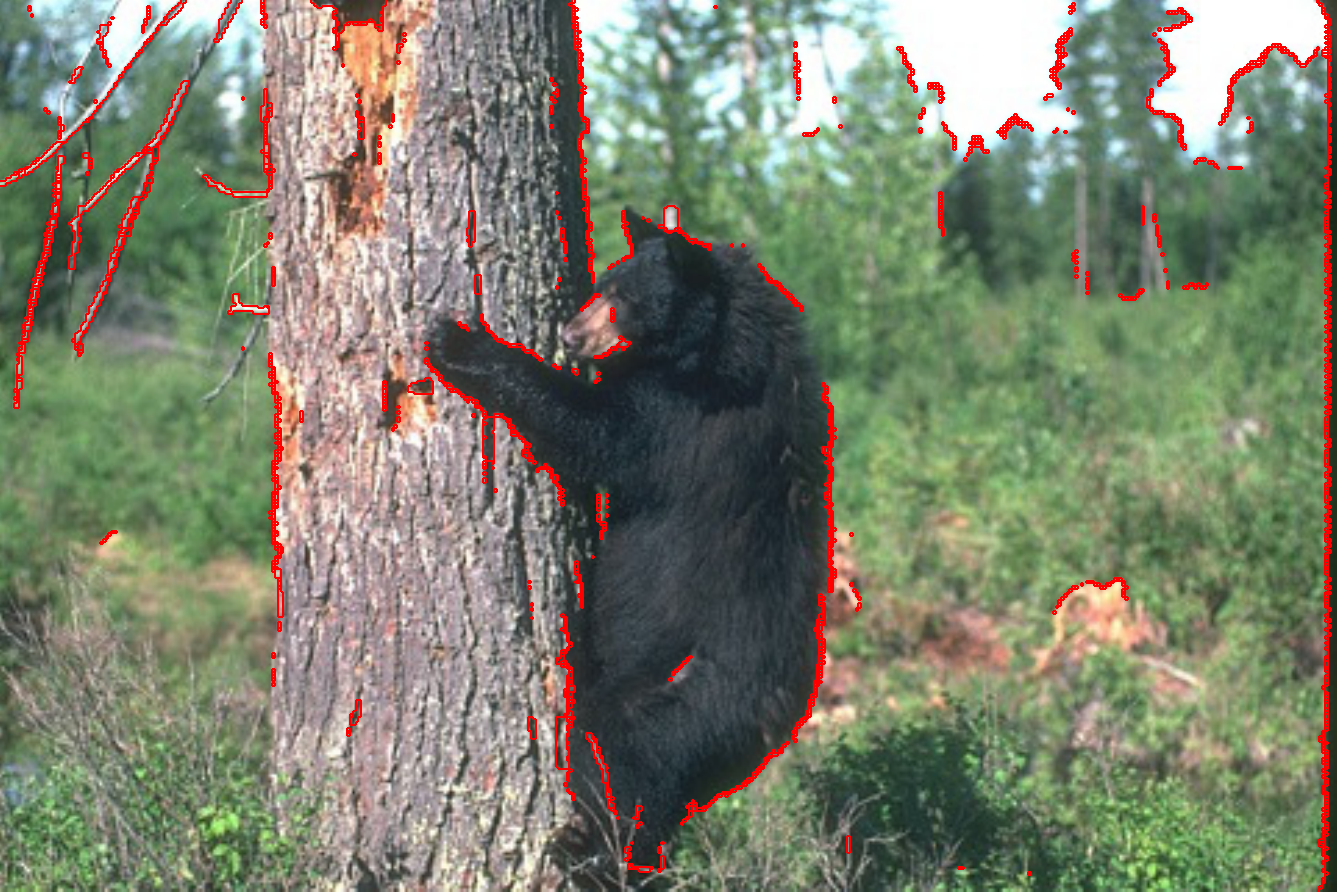}\,&
\includegraphics[width=0.195\textwidth]{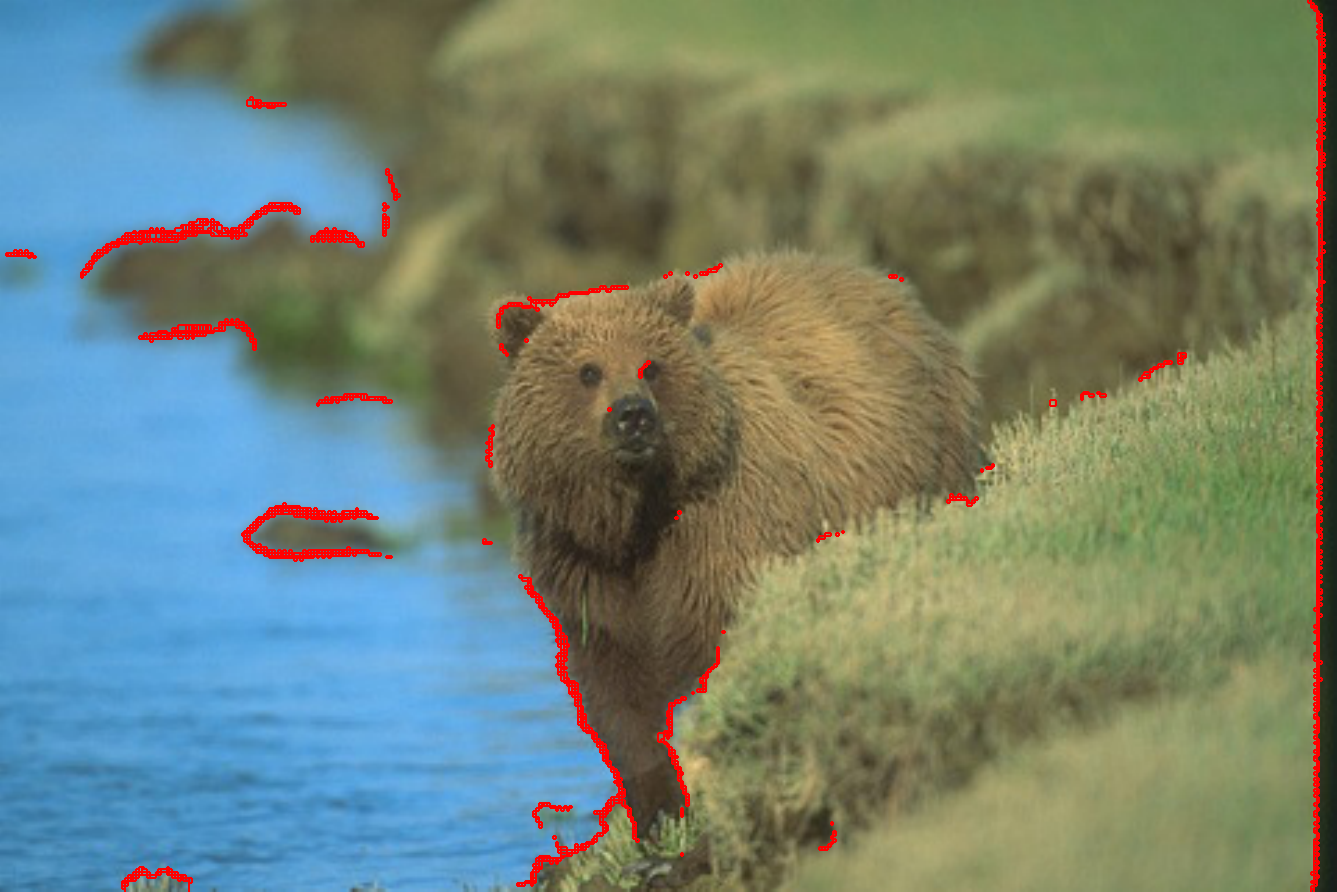}\,&
\includegraphics[width=0.195\textwidth]{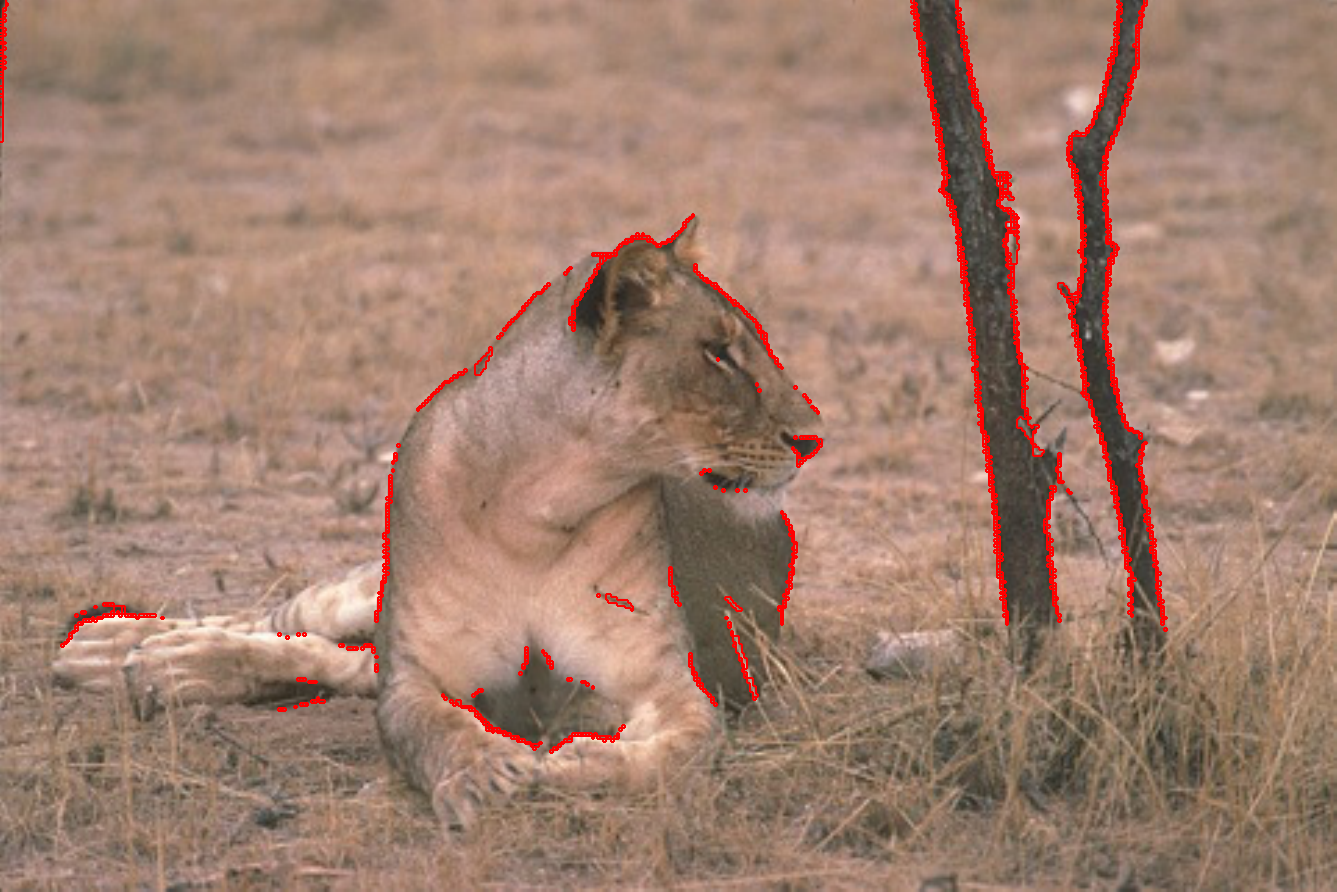}\,&
\includegraphics[width=0.195\textwidth]{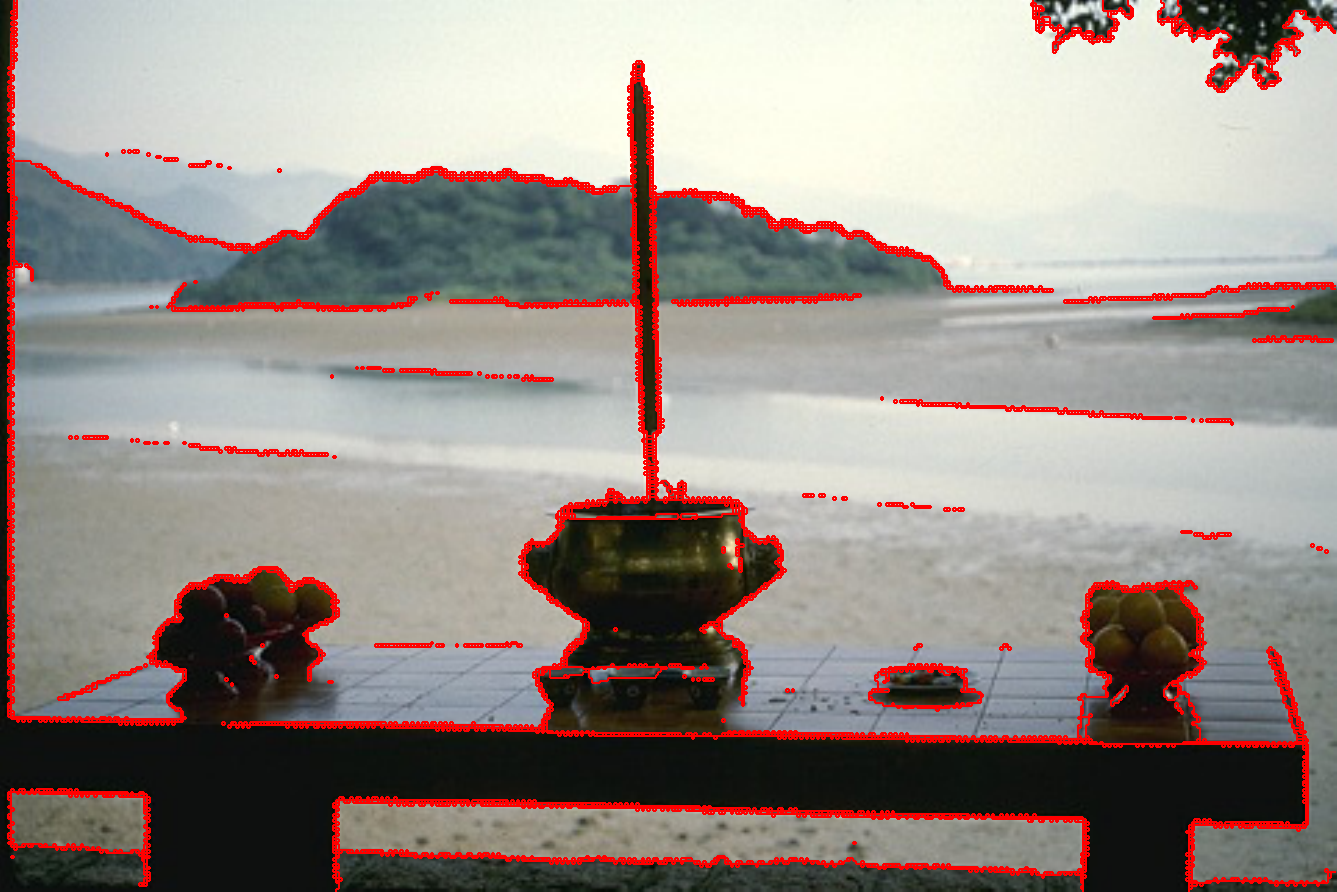}\,&
\includegraphics[width=0.195\textwidth]{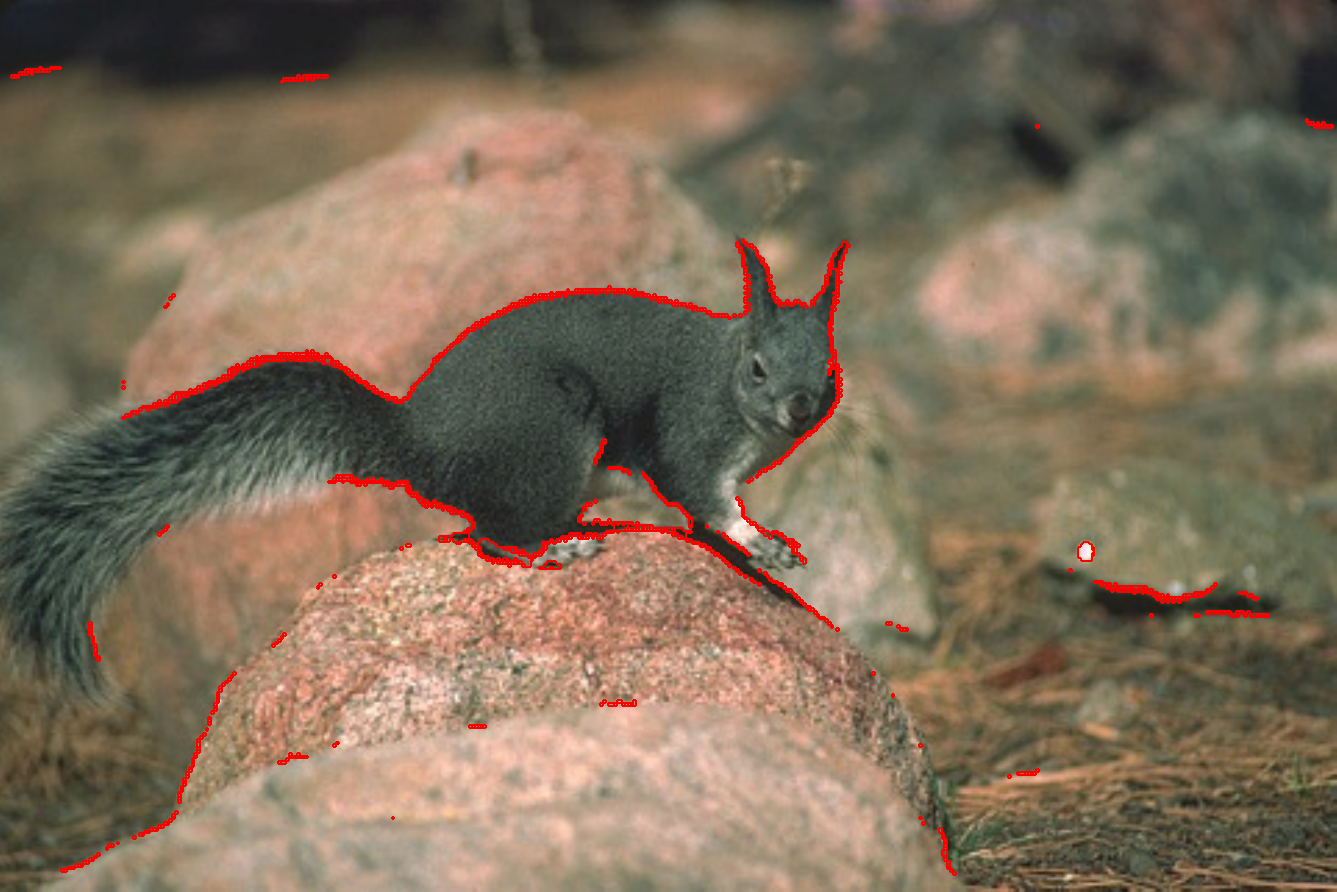}\\[0.1ex]
\includegraphics[width=0.195\textwidth]{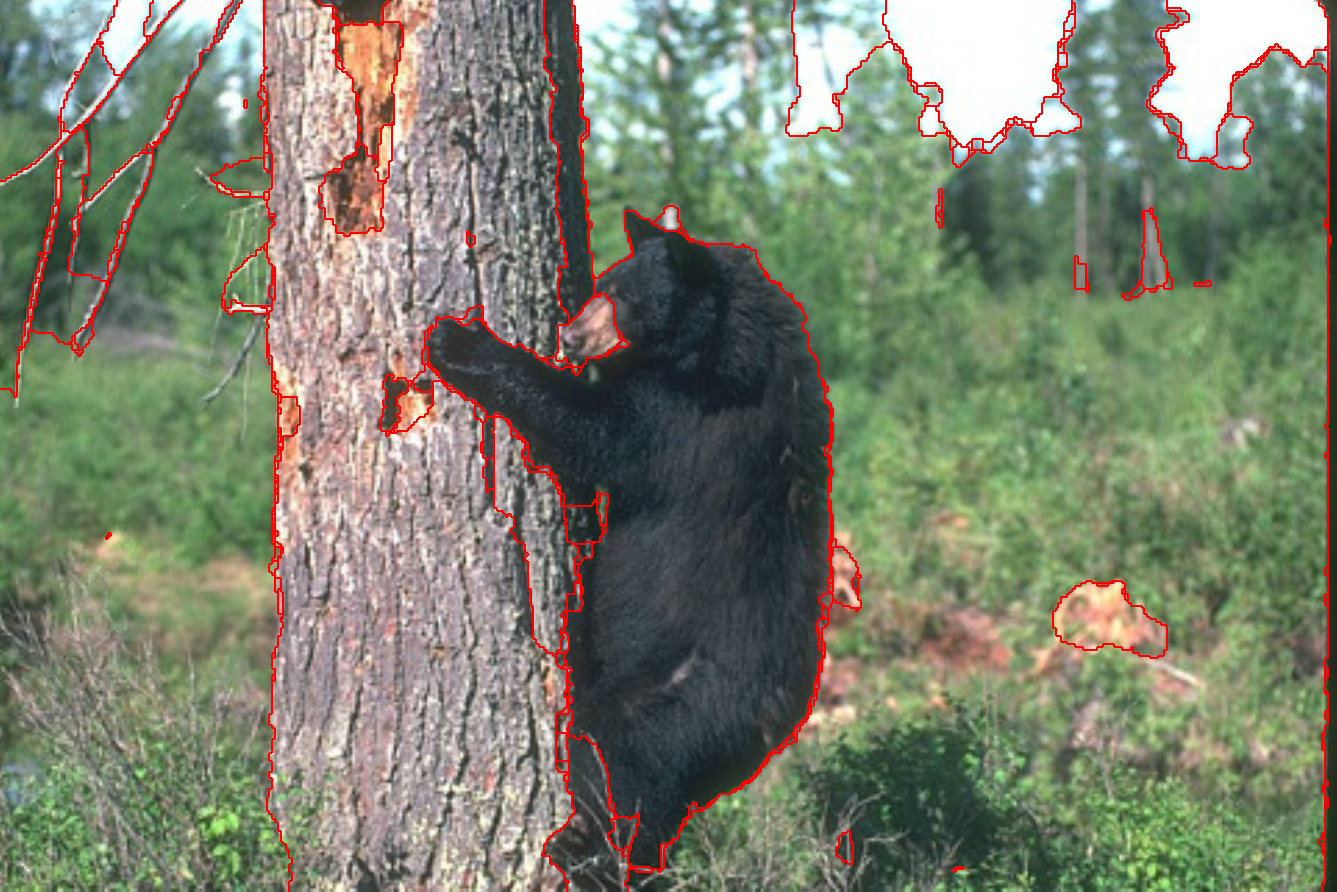}\,&
\includegraphics[width=0.195\textwidth]{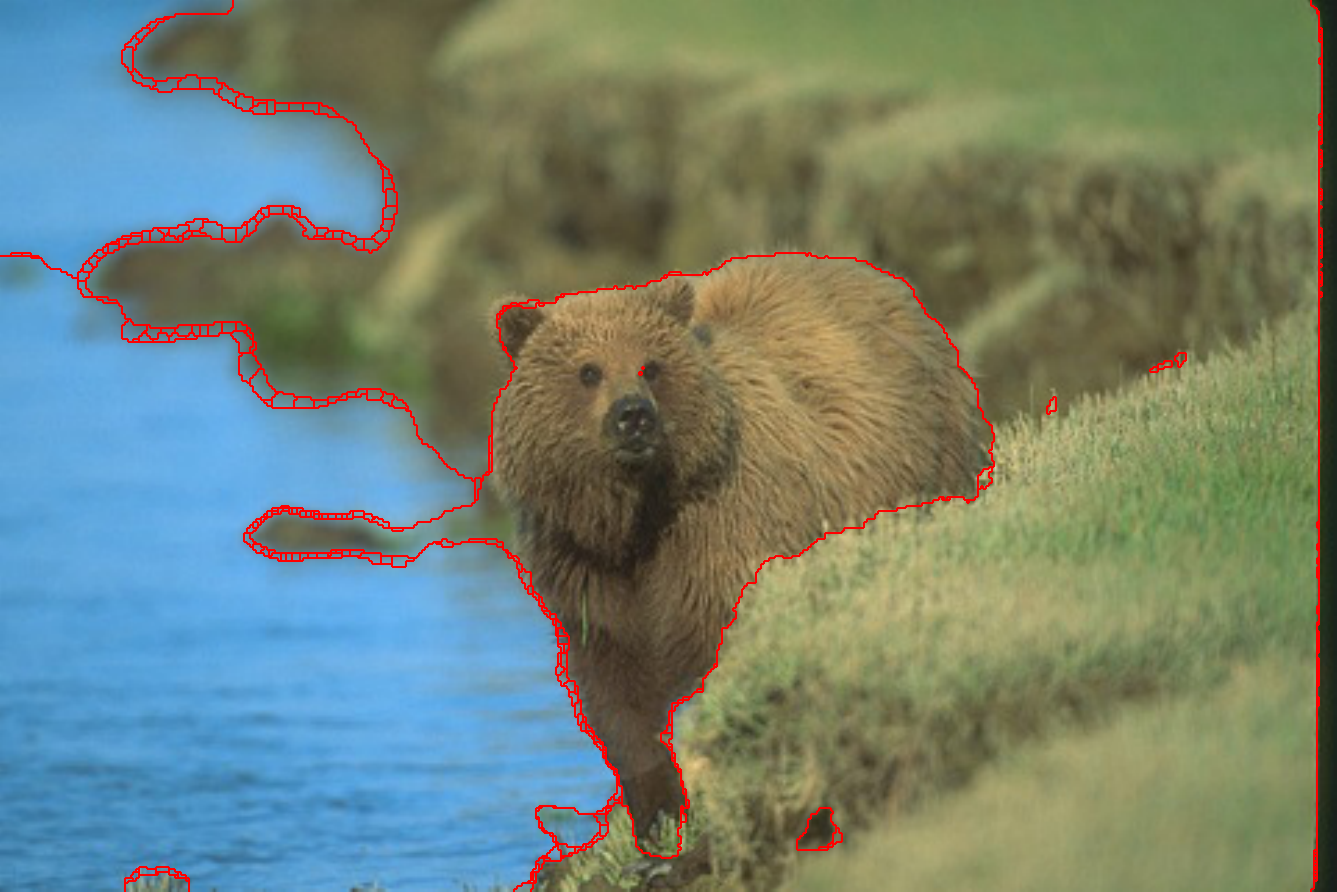}\,&
\includegraphics[width=0.195\textwidth]{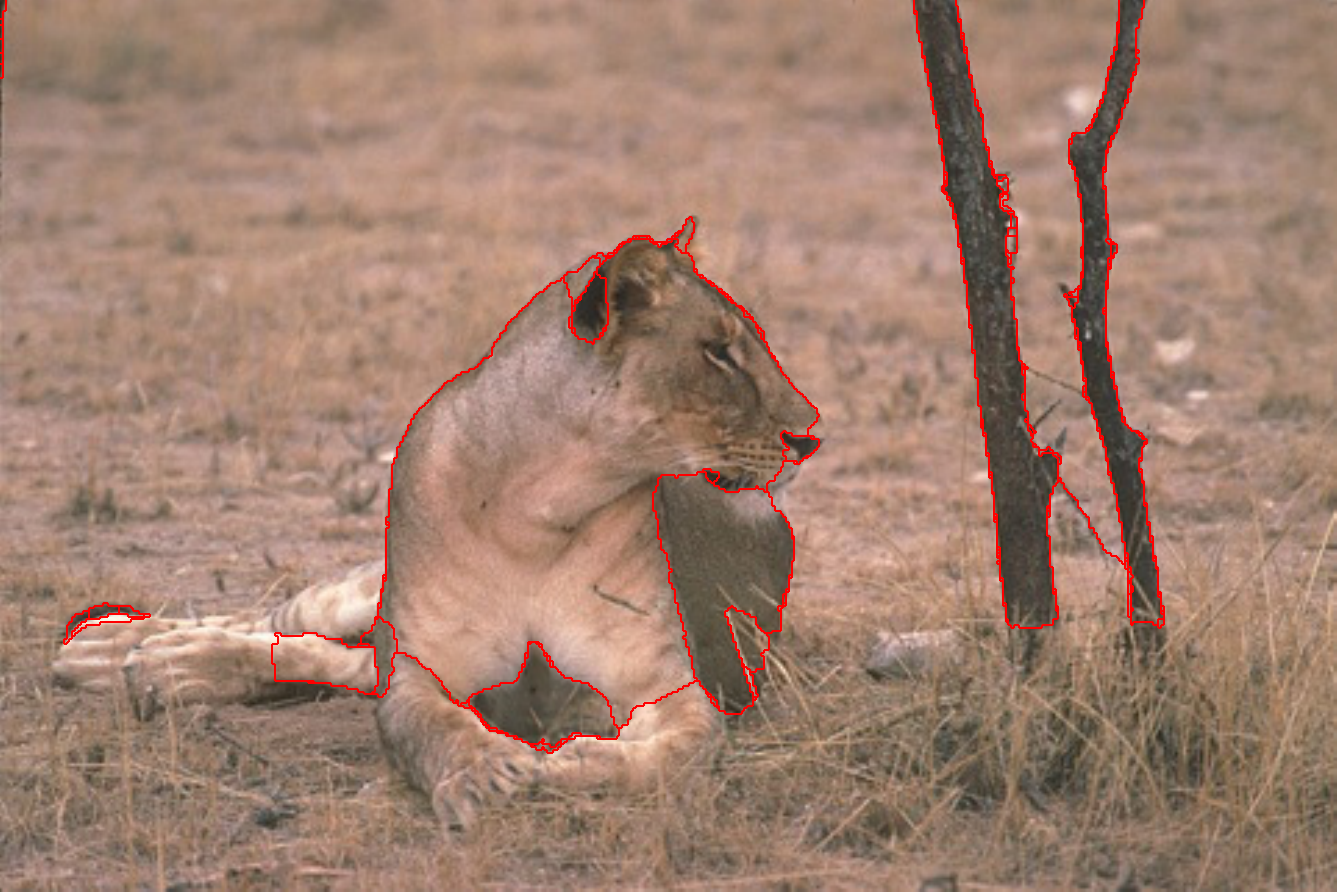}\,&
\includegraphics[width=0.195\textwidth]{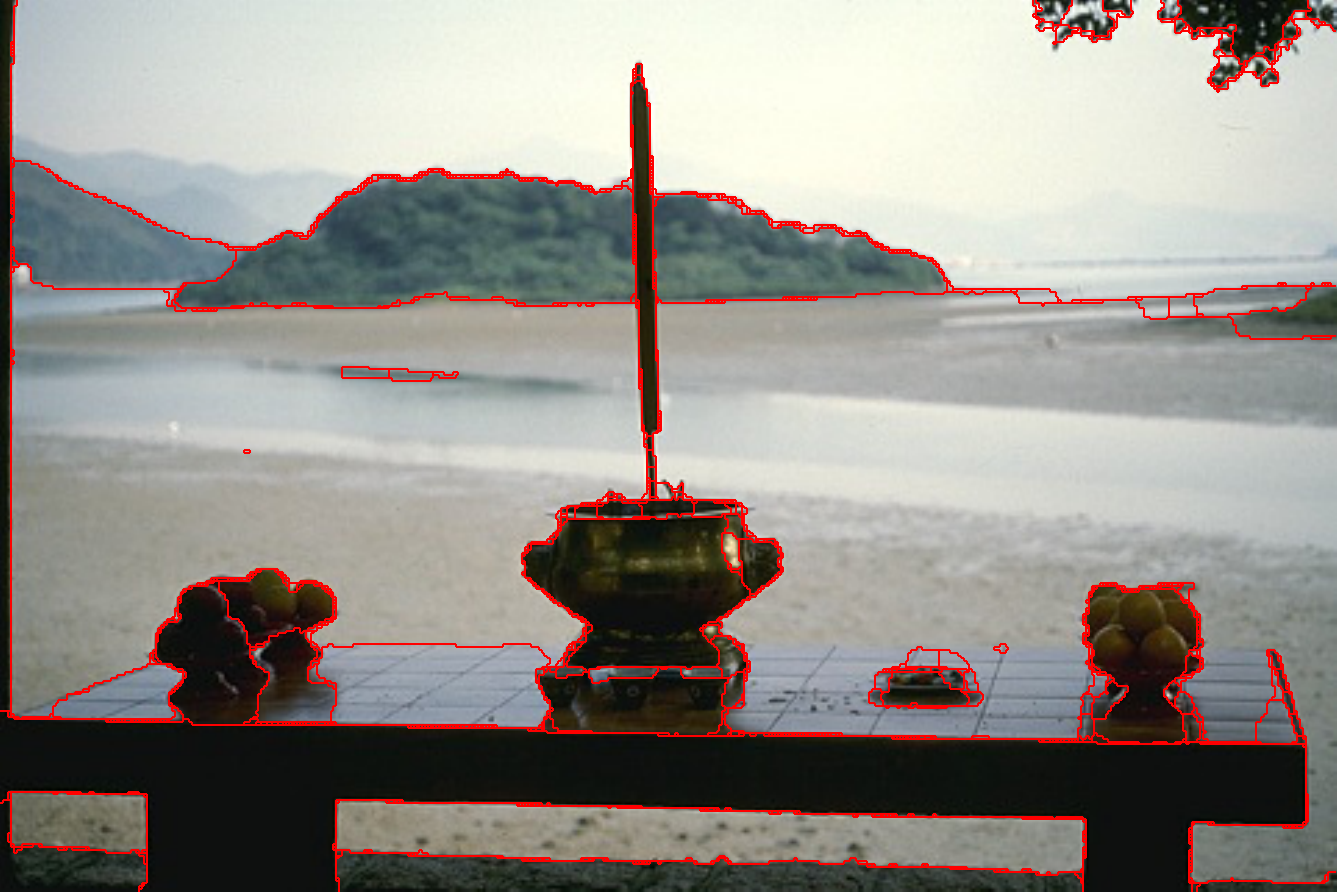}\,&
\includegraphics[width=0.195\textwidth]{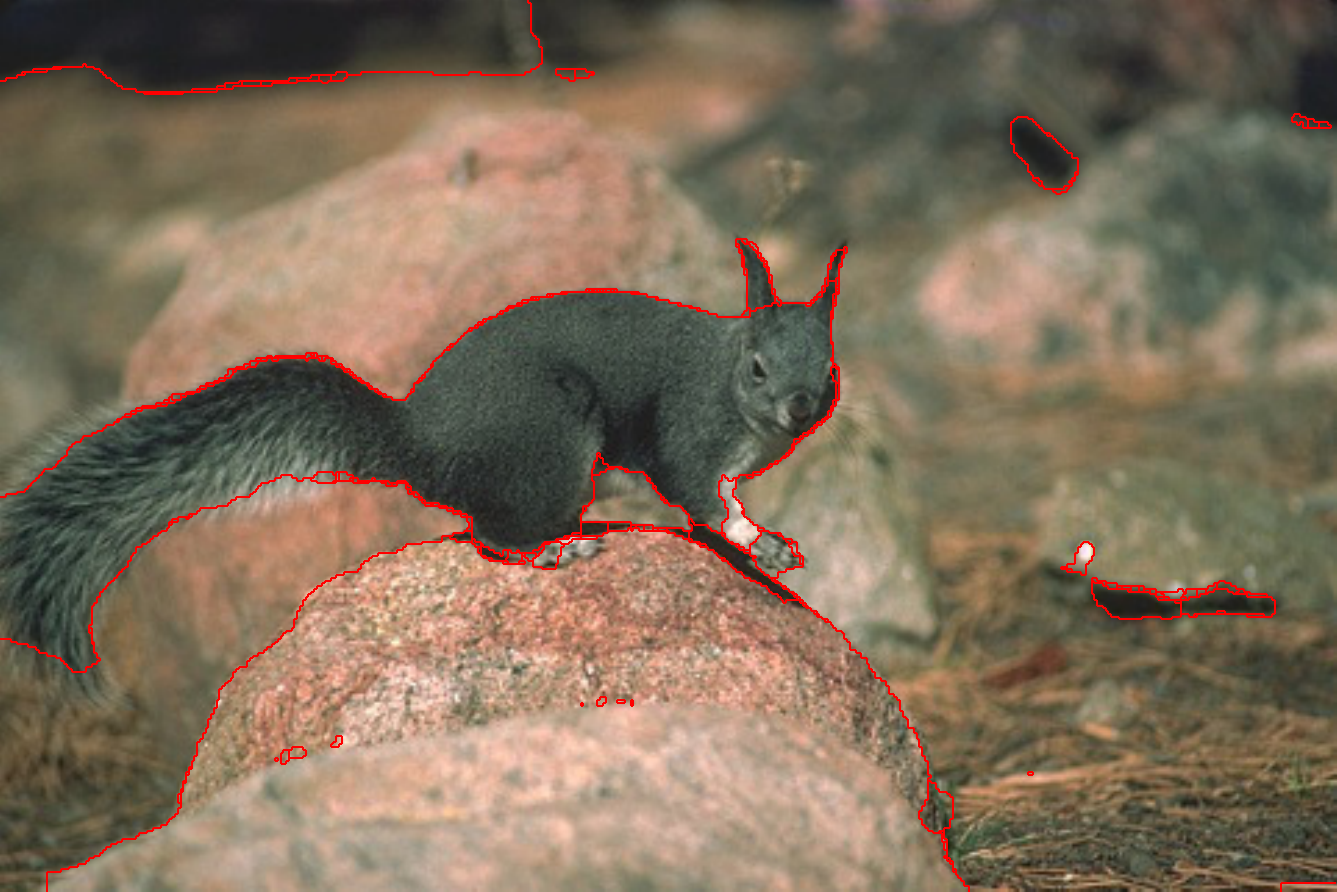}\\[-0.5ex]
\midrule
\includegraphics[width=0.195\textwidth]{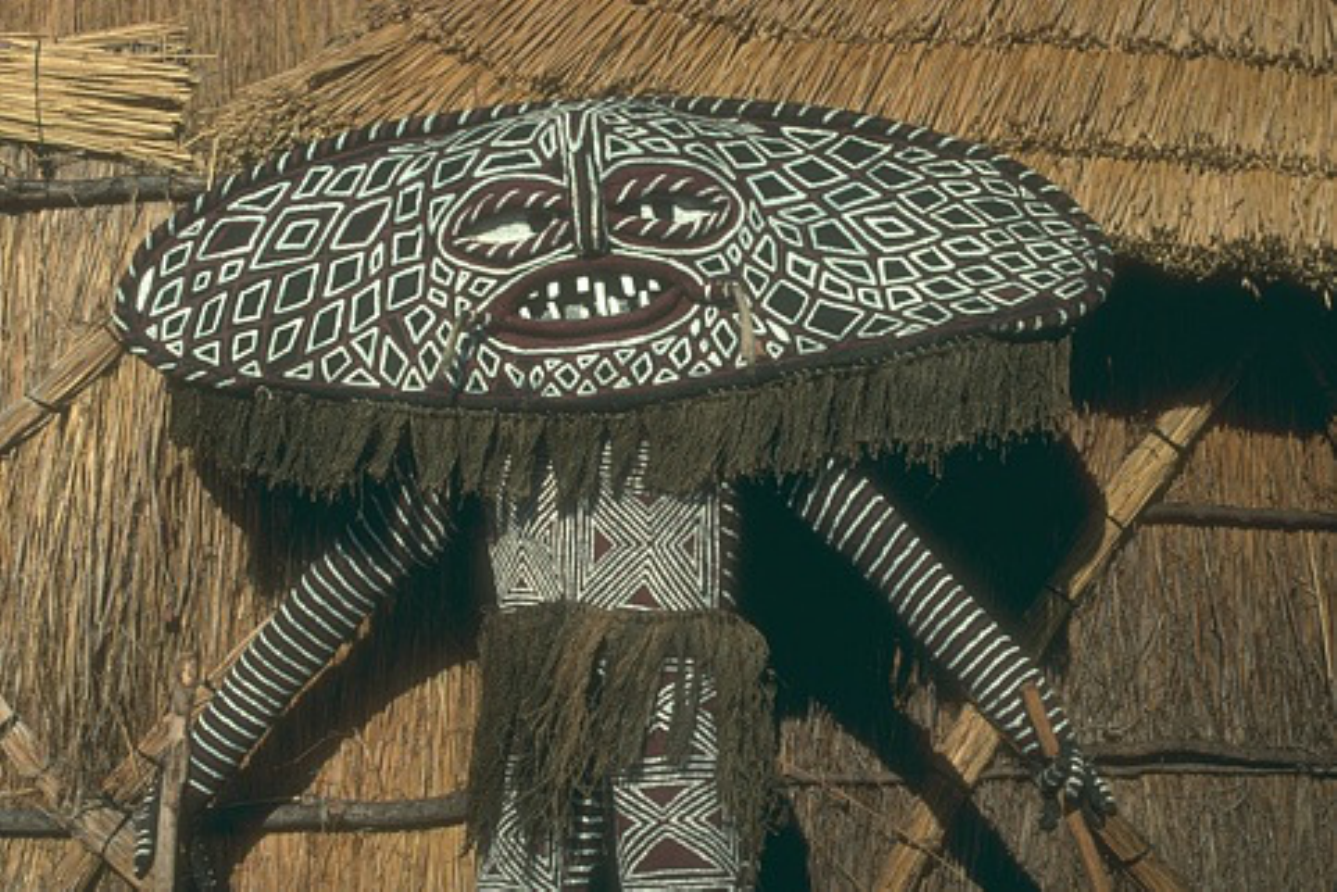}\,&
\includegraphics[width=0.195\textwidth]{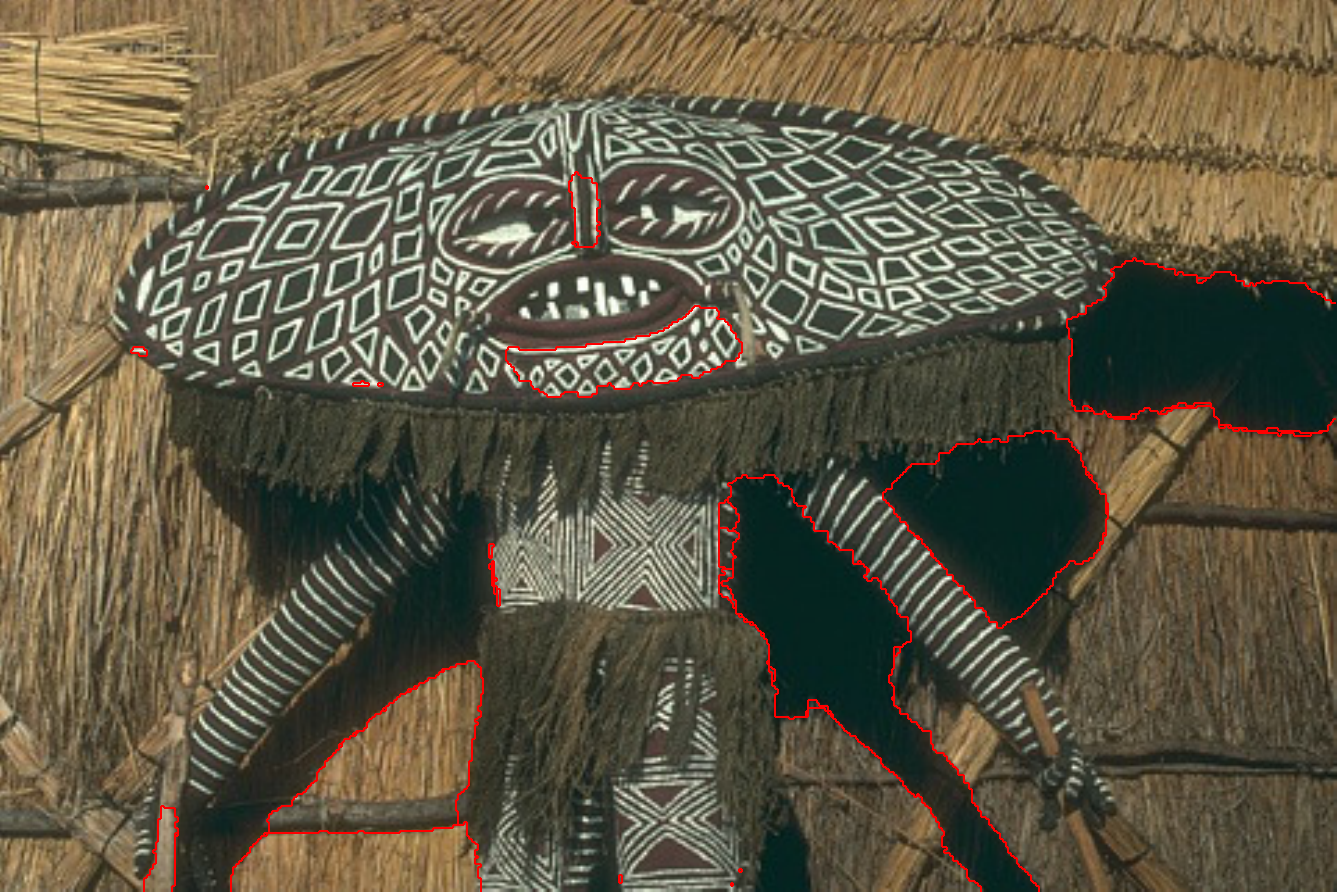}\,&
\includegraphics[width=0.195\textwidth]{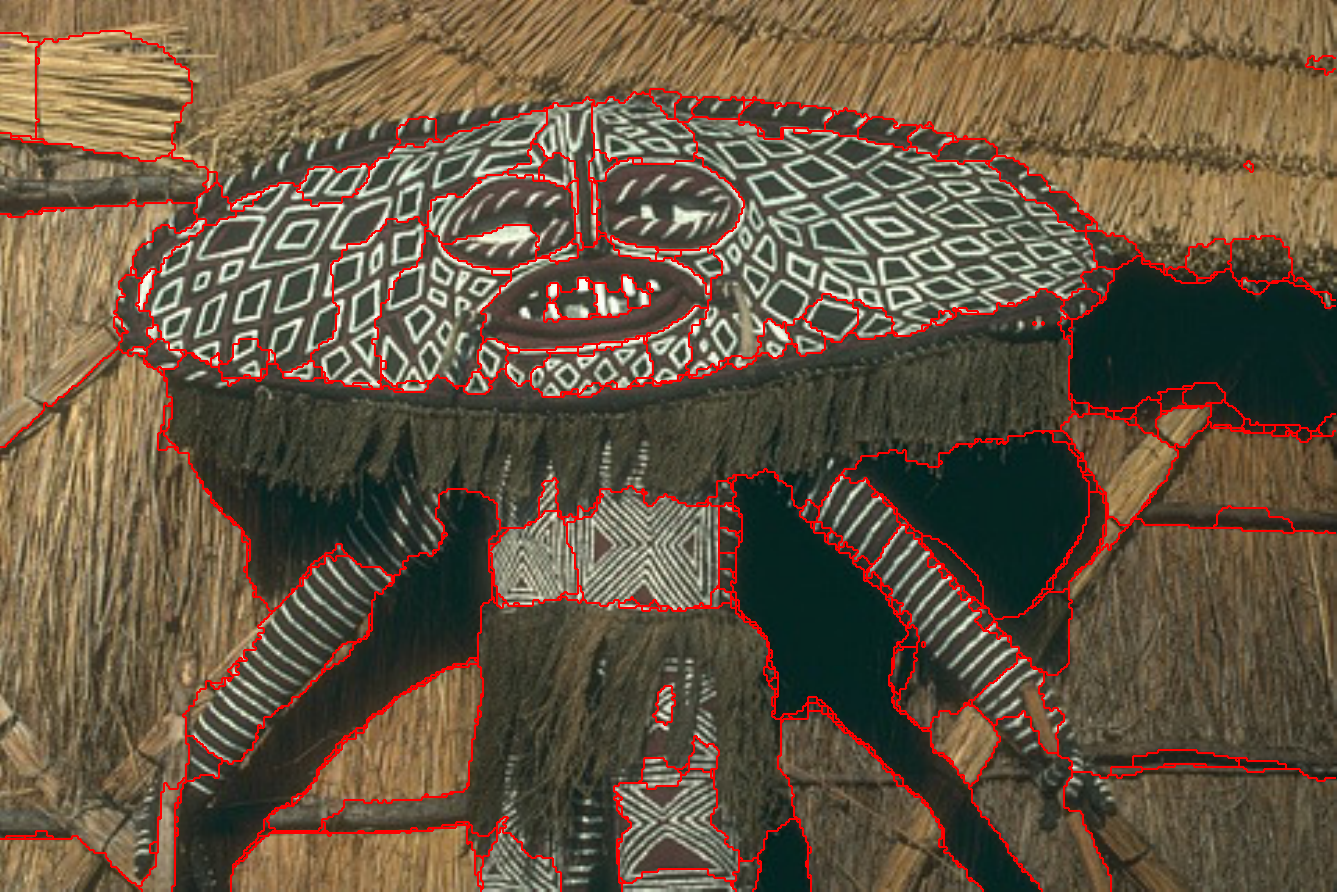}\,&
\includegraphics[width=0.195\textwidth]{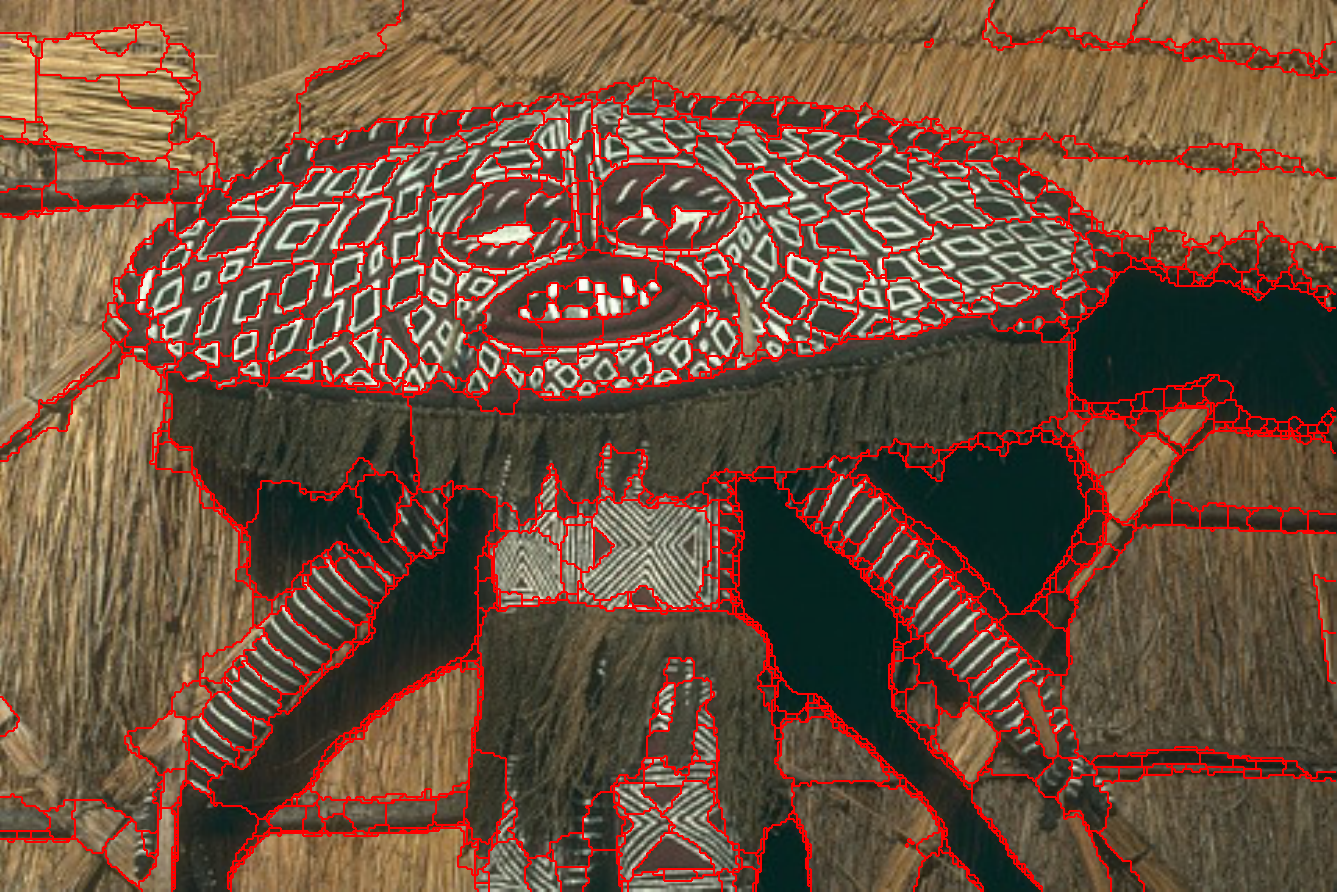}\,&
\includegraphics[width=0.195\textwidth]{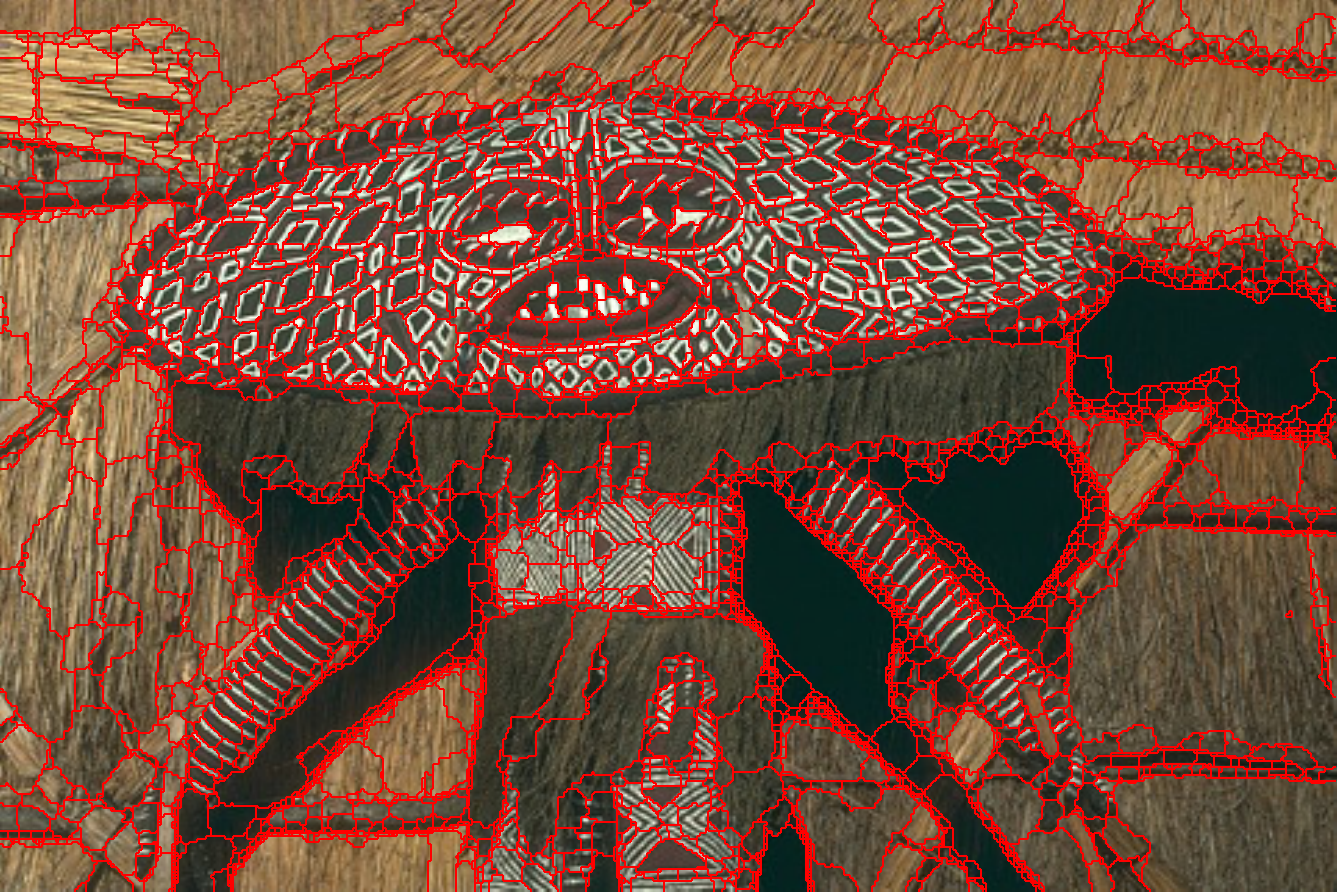}\,\\
$p^* = 0.3$ & $p^* = 0.4$ & $p^* = 0.5$ & $p^* = 0.6$ & $p^* = 0.7$\\
\end{tabular}
\caption{Depicted above is a comparison of 
decompositions defined by feasible solutions of the MP (\textbf{top} row for, the optimal $p^* = 0.8)$, with
decompositions defined by feasible solutions of the LMP (\textbf{middle} row, for the optimal $p^* = 0.5$ and $d^* = 20$). 
All solutions are found by 
Alg.~\ref{alg:KL_outer} (KLj), 
initialized with the output of 
Alg.~\ref{alg:AGG} (GAEC).
The decompositions defined by feasible solutions of the MP have closed contours but consist of tiny components on or near the boundary of desired components. This problem is overcome by feasible solutions of the LMP due to the long-range terms in the objective function.
Depicted in the \textbf{bottom} row is the effect of the prior probability $p^*$ which establishes a trade-off between over- and under-segmentation, reasonable in particular at $p^* = 0.5$.}
\label{fig:bias}
\end{figure*}

In contrast, feasible solutions of the LMP found by GAEC are improved effectively and efficiently by KLj. 
CGC is not practical for the larger, non-planar graphs of the instances of the LMP we define; the absolute runtime exceeds 48 hours for every image and $p^* = 0.5$.
Compared to the man-made decompositions in the benchmark,
feasible solutions of the LMP found by GAEC and improved by KLj are not significantly worse than the state of the art 
\cite{APBMM2014}
for this benchmark.
The effect of changing $p^*$ is shown for the average over all test images in
Fig.~\ref{figure:results}
and for one image in particular in 
Fig.~\ref{fig:bias}.
The best decompositions for this image as well as for all images on average are obtained for $p^* = 0.5$. 
The effect of changing $d^*$ is shown for the average over all test images in
Fig.~\ref{figure:algorithm-comparison} (on the left).
It can be seen from this figure that increasing $d^*$ from 5 to 10 improves results while further increasing $d^*$ to 20 does not change results noticably.

\begin{figure*}
\centering
\vspace{3ex}
\begin{tikzpicture}[font=\small]
\begin{axis}[
        width=0.67\columnwidth,
        height=0.67\columnwidth,
        xlabel={VI (false cut)\phantom{$^{3^3}$}}, 
        ylabel={VI (false join) },ylabel near ticks,
        xlabel near ticks,
        xmin=0, xmax=4,
        ymin=0, ymax=4,
        mark size=0.3ex,
        legend entries={{\bf MP}    GAEC-KLj,{\bf LMP}5  GAEC-KLj,{\bf LMP}10 GAEC-KLj,{\bf LMP}20 GAEC-KLj},
        legend style={draw=none,nodes={right}},
        legend pos=north east
    ]
    
    \addplot[forget plot, no marks, color=black!20] {-x+1};
    \addplot[forget plot, no marks, color=black!20] {-x+2};
    \addplot[forget plot, no marks, color=black!20] {-x+3};
    \addplot[forget plot, no marks, color=black!20] {-x+4};
    \addplot[forget plot, no marks, color=black!20] {-x+5};
    \addplot[forget plot, no marks, color=black!20] {-x+6};

    \addplot[
        forget plot, only marks, color=mycolor2!30,
        error bars/.cd, x dir=plus, x explicit, y dir=plus, y explicit,
    ] table [
        col sep=comma, 
        x index=6, y index=12,
        x error expr=\thisrowno{4}-\thisrowno{6}, y error expr=\thisrowno{10}-\thisrowno{12}
    ] {data-final/postICCVtestwta_KLold_keep_old_weights-eval_vipr.txt};

    \addplot[
        forget plot, only marks, color=mycolor2!30,
        error bars/.cd, x dir=minus, x explicit, y dir=minus, y explicit
    ] table [
        col sep=comma, 
        x index=6, y index=12,
        x error expr=-(\thisrowno{2}-\thisrowno{6}), y error expr=-(\thisrowno{8}-\thisrowno{12})
    ] {data-final/postICCVtestwta_KLold_keep_old_weights-eval_vipr.txt};

    \addplot[
        forget plot, only marks, color=mycolor3!30,
        error bars/.cd, x dir=plus, x explicit, y dir=plus, y explicit,
    ] table [
        col sep=comma, 
        x index=6, y index=12,
        x error expr=\thisrowno{4}-\thisrowno{6}, y error expr=\thisrowno{10}-\thisrowno{12}
    ] {data-final/postICCVtestwta5_KL_AGG_multicut_keep_old_weights-eval_vipr.txt};

    \addplot[
        forget plot, only marks, color=mycolor3!30,
        error bars/.cd, x dir=minus, x explicit, y dir=minus, y explicit
    ] table [
        col sep=comma, 
        x index=6, y index=12,
        x error expr=-(\thisrowno{2}-\thisrowno{6}), y error expr=-(\thisrowno{8}-\thisrowno{12})
    ] {data-final/postICCVtestwta5_KL_AGG_multicut_keep_old_weights-eval_vipr.txt};

    \addplot[
        forget plot, only marks, color=mycolor4!30,
        error bars/.cd, x dir=plus, x explicit, y dir=plus, y explicit,
    ] table [
        col sep=comma, 
        x index=6, y index=12,
        x error expr=\thisrowno{4}-\thisrowno{6}, y error expr=\thisrowno{10}-\thisrowno{12}
    ] {data-final/postICCVtestwta10_KL_AGG_multicut_keep_old_weights-eval_vipr.txt};

    \addplot[
        forget plot, only marks, color=mycolor4!30,
        error bars/.cd, x dir=minus, x explicit, y dir=minus, y explicit
    ] table [
        col sep=comma, 
        x index=6, y index=12,
        x error expr=-(\thisrowno{2}-\thisrowno{6}), y error expr=-(\thisrowno{8}-\thisrowno{12})
    ] {data-final/postICCVtestwta10_KL_AGG_multicut_keep_old_weights-eval_vipr.txt};

    \addplot[
        forget plot, only marks, color=mycolor1!30,
        error bars/.cd, x dir=plus, x explicit, y dir=plus, y explicit,
    ] table [
        col sep=comma, 
        x index=6, y index=12,
        x error expr=\thisrowno{4}-\thisrowno{6}, y error expr=\thisrowno{10}-\thisrowno{12}
    ] {data-final/postICCVtestwta20_KL_AGG_multicut_keep_old_weights-eval_vipr.txt};

    \addplot[
        forget plot, only marks, color=mycolor1!30,
        error bars/.cd, x dir=minus, x explicit, y dir=minus, y explicit
    ] table [
        col sep=comma, 
        x index=6, y index=12,
        x error expr=-(\thisrowno{2}-\thisrowno{6}), y error expr=-(\thisrowno{8}-\thisrowno{12})
    ] {data-final/postICCVtestwta20_KL_AGG_multicut_keep_old_weights-eval_vipr.txt};

    \addplot[
        only marks, color=mycolor2!30
    ] table [
        col sep=comma, 
        x index=6, y index=12
    ] {data-final/postICCVtestwta_KLold_keep_old_weights-eval_vipr.txt};
    
    \addplot[
        only marks, color=mycolor3
    ] table [
        col sep=comma, 
        x index=6, y index=12
    ] {data-final/postICCVtestwta5_KL_AGG_multicut_keep_old_weights-eval_vipr.txt};
    
    \addplot[
        only marks, color=mycolor4
    ] table [
        col sep=comma, 
        x index=6, y index=12
    ] {data-final/postICCVtestwta10_KL_AGG_multicut_keep_old_weights-eval_vipr.txt};
    
    \addplot[
        only marks, color=mycolor1
    ] table [
        col sep=comma, 
        x index=6, y index=12
    ] {data-final/postICCVtestwta20_KL_AGG_multicut_keep_old_weights-eval_vipr.txt};
\end{axis}
\end{tikzpicture}\hspace{0.6cm}
\begin{tikzpicture}[font=\small]
\begin{axis}[
        width=0.67\columnwidth,
        height=0.67\columnwidth,
        xlabel={Objective value\phantom{$^{3}$}}, 
        ylabel={Runtime [s]},ylabel near ticks,
        xlabel near ticks,
        ymode=log,
        ymin=0, ymax=100000,
        mark size=0.3ex,
        legend entries={KLj, KL ,CGC},
        legend style={draw=none,nodes={right}},
        legend pos=south west,
        clip marker paths=true
    ]
    
    \addplot[
        only marks, mark=+, color=mycolor1!30
    ] table [
        col sep=comma, 
        x index=1, y index=2
    ] {data-final/time-comp/all_kl_agg.txt};
    
    \addplot[
        only marks, mark=+, color=mycolor4!30
    ] table [
        col sep=comma, 
        x index=1, y index=2
    ] {data-final/time-comp/all_okl_agg.txt};
    
    \addplot[
        only marks, mark=+, color=mycolor3!30
    ] table [
        col sep=comma, 
        x index=1, y index=2
    ] {data-final/time-comp/all_cgc_agg.txt};

    \addplot[
        forget plot, only marks, color=mycolor1,
        error bars/.cd, x dir=plus, x explicit, y dir=plus, y explicit,
    ] table [
        col sep=comma, 
        x index=1, y index=4,
        x error expr=\thisrowno{3}-\thisrowno{1}, y error expr=\thisrowno{6}-\thisrowno{4}
    ] {data-final/time-comp/mean_kl_agg.txt};

    \addplot[
        forget plot, only marks, color=mycolor1,
        error bars/.cd, x dir=minus, x explicit, y dir=minus, y explicit
    ] table [
        col sep=comma, 
        x index=1, y index=4,
        x error expr=\thisrowno{1}-\thisrowno{2}, y error expr=\thisrowno{4}-\thisrowno{5}
    ] {data-final/time-comp/mean_kl_agg.txt};
    
    \addplot[
        forget plot, only marks, color=mycolor4,
        error bars/.cd, x dir=plus, x explicit, y dir=plus, y explicit,
    ] table [
        col sep=comma, 
        x index=1, y index=4,
        x error expr=\thisrowno{3}-\thisrowno{1}, y error expr=\thisrowno{6}-\thisrowno{4}
    ] {data-final/time-comp/mean_okl_agg.txt};

    \addplot[
        forget plot, only marks, color=mycolor4,
        error bars/.cd, x dir=minus, x explicit, y dir=minus, y explicit
    ] table [
        col sep=comma, 
        x index=1, y index=4,
        x error expr=\thisrowno{1}-\thisrowno{2}, y error expr=\thisrowno{4}-\thisrowno{5}
    ] {data-final/time-comp/mean_okl_agg.txt};
    
    \addplot[
        forget plot, only marks, color=mycolor3,
        error bars/.cd, x dir=plus, x explicit, y dir=plus, y explicit,
    ] table [
        col sep=comma, 
        x index=1, y index=4,
        x error expr=\thisrowno{3}-\thisrowno{1}, y error expr=\thisrowno{6}-\thisrowno{4}
    ] {data-final/time-comp/mean_cgc_agg.txt};

    \addplot[
        forget plot, only marks, color=mycolor3,
        error bars/.cd, x dir=minus, x explicit, y dir=minus, y explicit
    ] table [
        col sep=comma, 
        x index=1, y index=4,
        x error expr=\thisrowno{1}-\thisrowno{2}, y error expr=\thisrowno{4}-\thisrowno{5}
    ] {data-final/time-comp/mean_cgc_agg.txt};
    
    \addplot[
        only marks, color=mycolor1
    ] table [
        col sep=comma, 
        x index=1, y index=4
    ] {data-final/time-comp/mean_kl_agg.txt};
    
    \addplot[
        only marks, color=mycolor4
    ] table [
        col sep=comma, 
        x index=1, y index=4
    ] {data-final/time-comp/mean_okl_agg.txt};
    
    \addplot[
        only marks, color=mycolor3
    ] table [
        col sep=comma, 
        x index=1, y index=4
    ] {data-final/time-comp/mean_cgc_agg.txt};
\end{axis}
\end{tikzpicture}\hspace{0.9cm}
%
\begin{tikzpicture}[font=\small]
\begin{axis}[
        width=0.67\columnwidth,
        height=0.67\columnwidth,
        xlabel={Objective value + $2\cdot 10^3$}, 
        ylabel={Runtime [s]},ylabel near ticks,
        xlabel near ticks,
        ymode=log,
        xmode=log,
        ymin=0, ymax=100000,
        mark size=0.3ex,
        clip marker paths=true
    ]
    
    \addplot[
        only marks, mark=+, color=mycolor1!30
    ] table [
        col sep=comma, 
        x index=1, y index=2
    ] {data-final/time-comp/all_kl_tiles.txt};
    
    \addplot[
        only marks, mark=+, color=mycolor4!30
    ] table [
        col sep=comma, 
        x index=1, y index=2
    ] {data-final/time-comp/all_okl_tiles.txt};
    
    \addplot[
        only marks, mark=+, color=mycolor3!30
    ] table [
        col sep=comma, 
        x index=1, y index=2
    ] {data-final/time-comp/all_cgc_tiles.txt};

    \addplot[
        forget plot, only marks, color=mycolor1,
        error bars/.cd, x dir=plus, x explicit, y dir=plus, y explicit,
    ] table [
        col sep=comma, 
        x index=1, y index=4,
        x error expr=\thisrowno{3}-\thisrowno{1}, y error expr=\thisrowno{6}-\thisrowno{4}
    ] {data-final/time-comp/mean_kl_tiles.txt};

    \addplot[
        forget plot, only marks, color=mycolor1,
        error bars/.cd, x dir=minus, x explicit, y dir=minus, y explicit
    ] table [
        col sep=comma, 
        x index=1, y index=4,
        x error expr=\thisrowno{1}-\thisrowno{2}, y error expr=\thisrowno{4}-\thisrowno{5}
    ] {data-final/time-comp/mean_kl_tiles.txt};
    
    \addplot[
        forget plot, only marks, color=mycolor4,
        error bars/.cd, x dir=plus, x explicit, y dir=plus, y explicit,
    ] table [
        col sep=comma, 
        x index=1, y index=4,
        x error expr=\thisrowno{3}-\thisrowno{1}, y error expr=\thisrowno{6}-\thisrowno{4}
    ] {data-final/time-comp/mean_okl_tiles.txt};

    \addplot[
        forget plot, only marks, color=mycolor4,
        error bars/.cd, x dir=minus, x explicit, y dir=minus, y explicit
    ] table [
        col sep=comma, 
        x index=1, y index=4,
        x error expr=\thisrowno{1}-\thisrowno{2}, y error expr=\thisrowno{4}-\thisrowno{5}
    ] {data-final/time-comp/mean_okl_tiles.txt};
    
    \addplot[
        forget plot, only marks, color=mycolor3,
        error bars/.cd, x dir=plus, x explicit, y dir=plus, y explicit,
    ] table [
        col sep=comma, 
        x index=1, y index=4,
        x error expr=\thisrowno{3}-\thisrowno{1}, y error expr=\thisrowno{6}-\thisrowno{4}
    ] {data-final/time-comp/mean_cgc_tiles.txt};

    \addplot[
        forget plot, only marks, color=mycolor3,
        error bars/.cd, x dir=minus, x explicit, y dir=minus, y explicit
    ] table [
        col sep=comma, 
        x index=1, y index=4,
        x error expr=\thisrowno{1}-\thisrowno{2}, y error expr=\thisrowno{4}-\thisrowno{5}
    ] {data-final/time-comp/mean_cgc_tiles.txt};
    
    \addplot[
        only marks, color=mycolor1
    ] table [
        col sep=comma, 
        x index=1, y index=4
    ] {data-final/time-comp/mean_kl_tiles.txt};
    
    \addplot[
        only marks, color=mycolor4
    ] table [
        col sep=comma, 
        x index=1, y index=4
    ] {data-final/time-comp/mean_okl_tiles.txt};
    
    \addplot[
        only marks, color=mycolor3
    ] table [
        col sep=comma, 
        x index=1, y index=4
    ] {data-final/time-comp/mean_cgc_tiles.txt};
\end{axis}
\end{tikzpicture}
\caption{Depicted on the \textbf{left} is the effect of the lifting distance $d^*$.
It can be seen that increasing $d^*$ from 5 to 10 improves the quality of image decompositions as measured by the VI; further increasing $d^*$ to 20 does not result in a measurable improvement.
In the \textbf{middle}, a comparison of Alg.~\ref{alg:KL_outer} (KLj)
with CGC
\cite{beier-2014}
and the implementation of the Kernighan-Lin Algorithm in
\cite{andres-2012-opengm,kappes-2015} (KL) is given.
Every point corresponds to one instance of the MP ($p^* = 0.5$) defined w.r.t.~one test image in the BSDS-500 benchmark
\cite{arbelaez-2011}.
All algorithms are initialized here with the output of Alg.~\ref{alg:AGG} (GAEC) for this instance.
On the \textbf{right},
analogous results are shown for all algorithms initialized with a decomposition of the pixel grid into tiles of $30 \cdot 30$ pixels.
It can be seen that KLj strikes a favorable balance between objective value and runtime.}
\label{figure:algorithm-comparison}
\end{figure*}
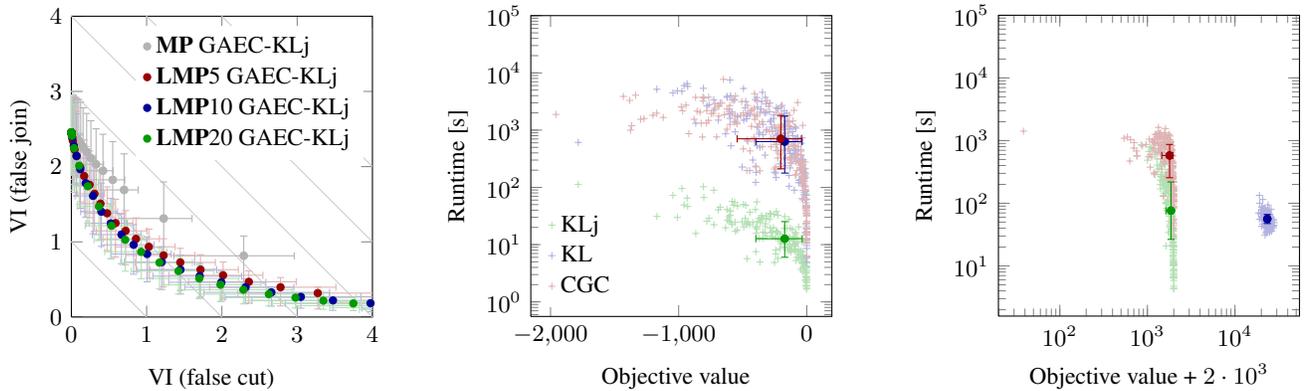
\begin{table}
\centering
\footnotesize
\begin{tabular}{lllll}
\toprule
& Boundary & Volume\\
\midrule
& F-measure & Covering & RI & VI \cite{meila-2007}\\
\midrule
gPb-owt-ucm \cite{arbelaez-2011} &0.73&0.59&0.83&1.69\\
SE+MS+SH\cite{DollarICCV13edges}+ucm & 0.73& 0.59 & 0.83 & 1.71  \\
SE+multi+ucm \cite{APBMM2014} &0.75 &0.61&0.83&1.57 \\
\midrule
SE+\textbf{MP} GAEC & 0.71&0.50& 0.80 &2.36\\
SE+\textbf{MP} GAEC-KLj & 0.71&0.50& 0.80 &2.36\\
SE+\textbf{MP} 1-KLj & 0.71 & 0.49 & 0.80 & 2.41\\
SE+\textbf{MP} GAEC-CGC & 0.71&0.50& 0.80 &2.23\\
\midrule
SE+\textbf{LMP}10 GAEC & 0.71&0.51& 0.80 &2.33\\
SE+\textbf{LMP}10 GAEC-KLj & 0.73&0.58& 0.82 &1.76\\
SE+\textbf{LMP}10 1-KLj & 0.73 & 0.58 & 0.82 & 1.76\\
\midrule
SE+\textbf{LMP}20 GAEC & 0.71&0.52& 0.80 &2.22\\
SE+\textbf{LMP}20 GAEC-KLj & 0.73&0.58& 0.82 &1.74\\
SE+\textbf{LMP}20 1-KLj &0.73 & 0.57 & 0.82 & 1.75\\
\bottomrule
\end{tabular}\\[1.5ex]
\caption{Written above are boundary and volume metrics 
measuring the distance between the man-made decompositions of the BSDS-500 benchmark
\cite{arbelaez-2011}
and the decompositions defined by multicuts (MP), lifted multicuts (LMP) and top-performing competing methods
\cite{arbelaez-2011,APBMM2014,DollarICCV13edges}.
Parameters are fixed for the entire data set (ODS).}
\label{table:results}
\end{table}

\subsection{Mesh Decomposition}

\begin{table}
\scriptsize
\centering
\begin{tabular}{@{}l@{\hspace{2.5ex}}l@{\hspace{2.5ex}}l@{\hspace{2.5ex}}l@{\hspace{2.5ex}}l@{\hspace{2.5ex}}l@{\hspace{2.5ex}}l@{\hspace{2.5ex}}l@{\hspace{2.5ex}}l@{\hspace{2.5ex}}l@{\hspace{2.5ex}}l@{}}
\toprule
& \cite{Zhanga} 
& \cite{Kin-ChungAu2011} 
& \cite{Kalogerakis2010}
& & \multicolumn{2}{@{}l@{}}{\textbf{MP}} 
& \multicolumn{2}{@{}l@{}}{\textbf{LMP}70} 
& \multicolumn{2}{@{}l@{}}{\textbf{LMP}opt}\\
\midrule
      &        RI    & RI    & RI    & VI    & RI    & VI & RI    & VI    & RI    & VI \\
Human &        0.89  & 0.88  & 0.88  & 1.43  &   0.21 & 2.93 &    0.86  & 1.62  & 0.87  & 1.79 \\
Cup &        0.80  & 0.79  & 0.90  & 0.35  & 0.74 & 0.68 &   0.89  & 0.39  & 0.90  & 0.39 \\
Glasses &       0.91  & 0.90  & 0.86  & 0.68  & 0.35 & 1.83 &   0.84  & 0.76  & 0.90  & 0.68 \\
Airplane &        0.89  & 0.87  & 0.92  & 0.67  &  0.69 & 1.39 &   0.92  & 0.82  & 0.92  & 0.83 \\
Ant &        0.98  & 0.96  & 0.98  & 0.37  & 0.93 & 0.67 &      0.98  & 0.42  & 0.98  & 0.42 \\
Chair &        0.89  & 0.88  & 0.95  & 0.43  & 0.79 & 0.98 &   0.93  & 0.55  & 0.93  & 0.55 \\
Octopus &        0.98  & 0.96  & 0.98  & 0.29  &  0.86 & 0.80 &    0.98  & 0.35  & 0.98  & 0.33 \\
Table &        0.90  & 0.94  & 0.94  & 0.28  &  0.76 & 0.81 &  0.94  & 0.28  & 0.94  & 0.29 \\
Teddy &        0.97  & 0.95  & 0.97  & 0.37  &  0.69 & 1.37 &  0.96  & 0.51  & 0.96  & 0.50 \\
Hand &        0.92  & 0.89  & 0.90  & 0.85  &  0.29 & 2.36 &     0.83  & 1.26  & 0.85  & 1.32 \\
Plier &        0.91  & 0.93  & 0.95  & 0.57  &  0.25 & 2.14 &   0.91  & 0.88  & 0.93  & 0.84 \\
Fish &        0.70  & 0.76  & 0.87  & 0.70  &  0.64 & 1.27&     0.80  & 1.09  & 0.80  & 1.09 \\
Bird &        0.91  & 0.90  & 0.91  & 0.73  &  0.67 & 1.36 &   0.93  & 0.88  & 0.93  & 0.99 \\
Armadillo &        0.91  & 0.89  & 0.93  & 1.11  & 0.21 & 3.27 &     0.92  & 1.60  & 0.92  & 1.48 \\
Bust &        0.75  & 0.76  & 0.76  & 1.35  &  0.42 & 1.67 &     0.69  & 2.25  & 0.69  & 2.25 \\
Mech &        0.87  & 0.88  & 0.89  & 0.46  &  0.78 & 0.69 &    0.84  & 0.59  & 0.84  & 0.59 \\
Bearing &        0.83  & 0.82  & 0.91  & 0.45  & 0.87 & 0.60 &     0.84  & 0.69  & 0.84  & 0.69 \\
Vase &        0.88  & 0.83  & 0.85  & 0.75  & 0.55 & 1.34 &   0.83  & 0.90  & 0.84  & 0.87 \\
FourLeg &        0.86  & 0.82  & 0.86  & 1.34  & 0.30 & 2.58 &  0.84  & 1.84  & 0.84  & 1.72 \\
\midrule
Average &        0.88  & 0.87  & 0.91  & 0.69  & 0.58 & 1.51 &  0.88  & 0.93  & 0.89  & 0.93 \\
\bottomrule
\end{tabular}\\[1.5ex]
\caption{Written above are boundary and volume metrics 
measuring the distance between the man-made decompositions of meshes in the 
Princeton Benchmark
and the decompositions defined by multicuts (MP), lifted multicuts (LMP) and top-performing competing methods
\cite{Zhanga,Kin-ChungAu2011,Kalogerakis2010}.
The evaluation is for a fixed parameter set for the entire database ($p^* = 0.55$, $d^* = 70$: LMP70)
as well as for the best parameter set we found for each class of meshes (LMPopt). 
Results for the MP are for the optimal $p^* = 0.9$.}
\label{tab:mesh}
\end{table}

\begin{figure}
\centering
\includegraphics[width=\columnwidth]{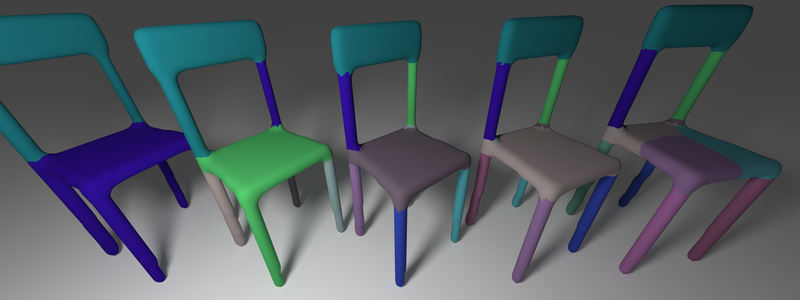}
\caption{Depicted above is the effect of varying the prior probability $p^*$ of adjacent triangles being in distinct components. Here, $p^* \in \{0.5, 0.55, 0.58, 0.6, 0.62\}$, from left to right.}
\label{fig:meshbias}
\end{figure}

\begin{figure*}[t]
\includegraphics[width=0.195\textwidth]{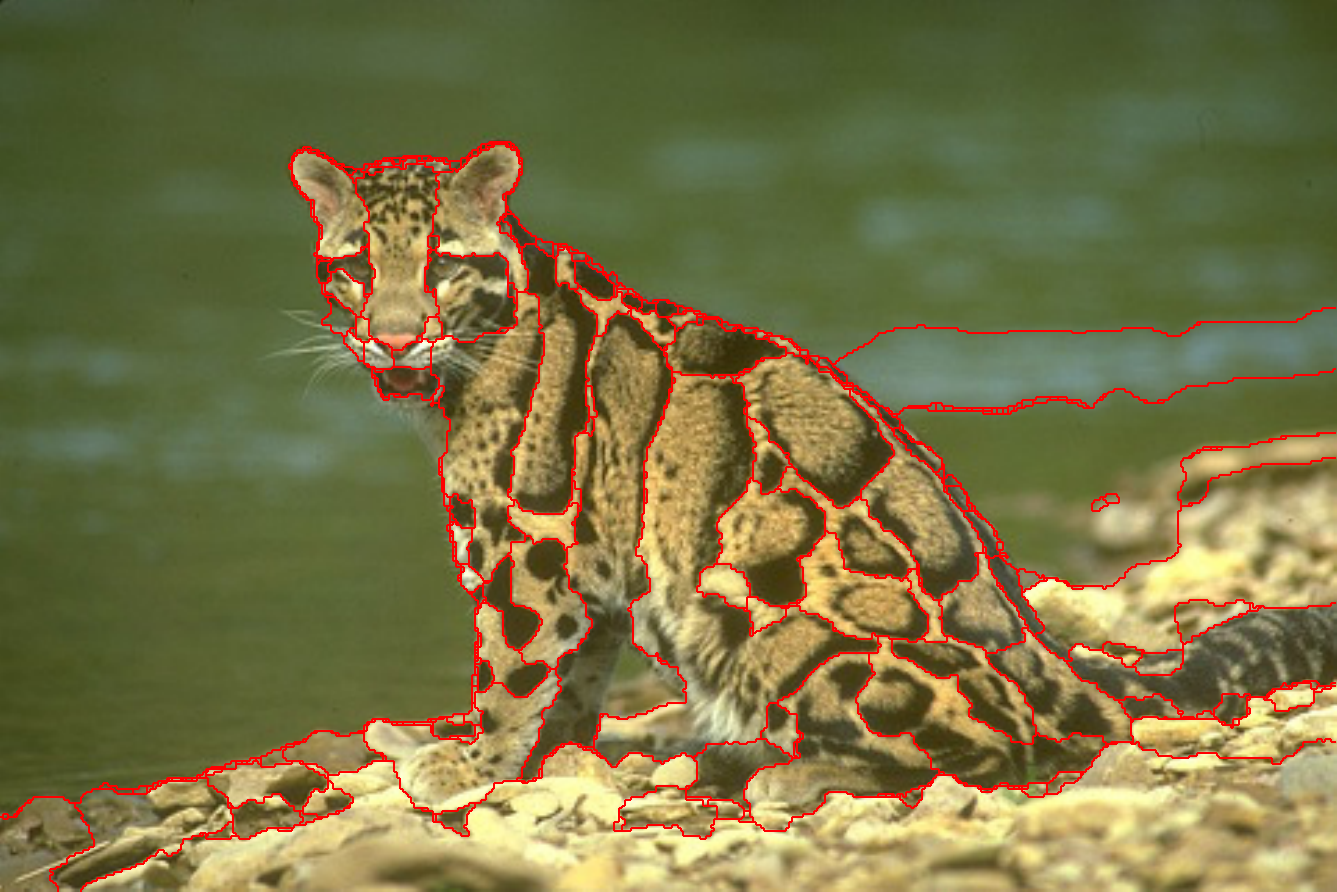}\hfill
\includegraphics[width=0.195\textwidth]{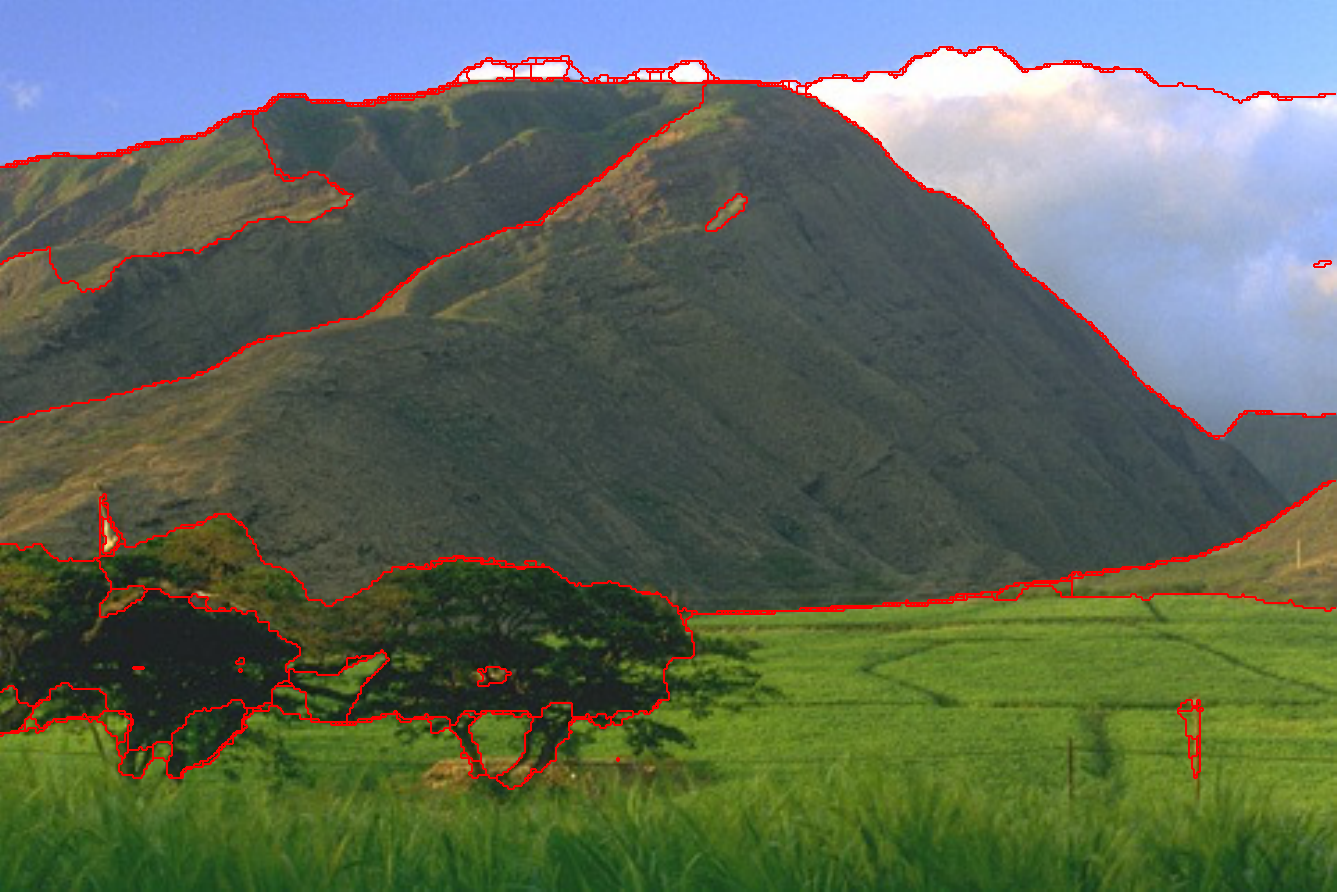}\hfill
\includegraphics[width=0.195\textwidth]{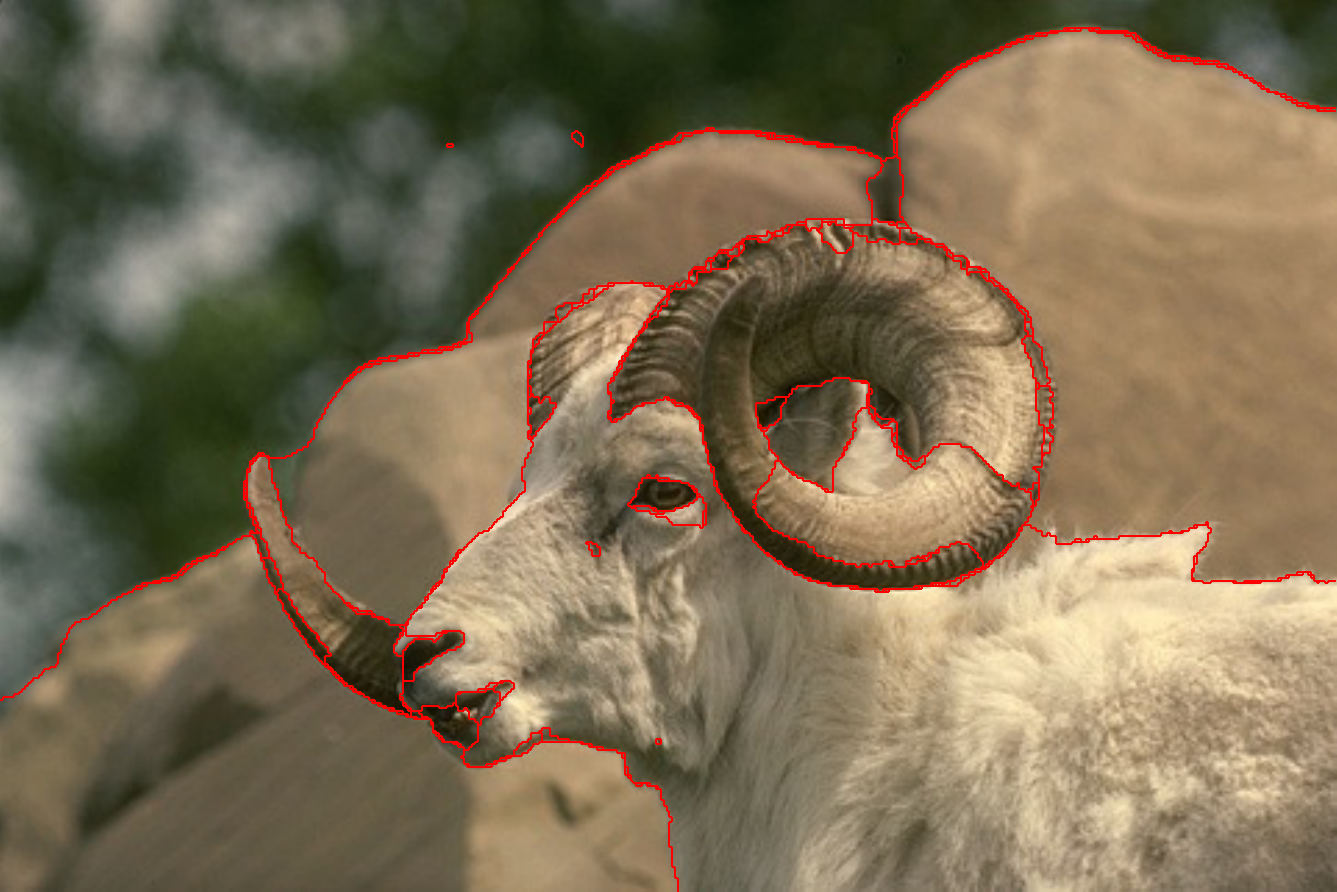}\hfill
\includegraphics[width=0.195\textwidth]{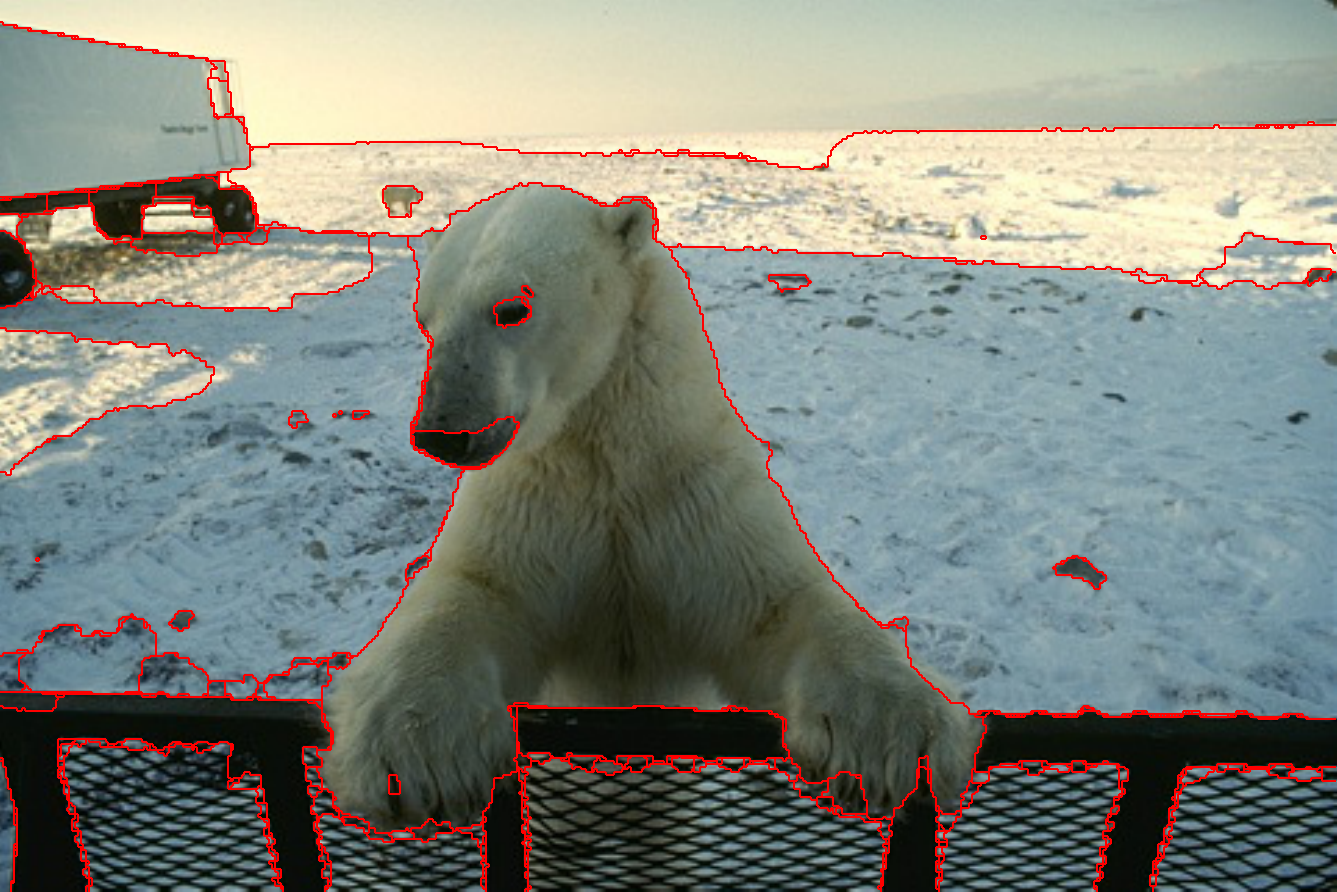}\hfill
\includegraphics[width=0.195\textwidth]{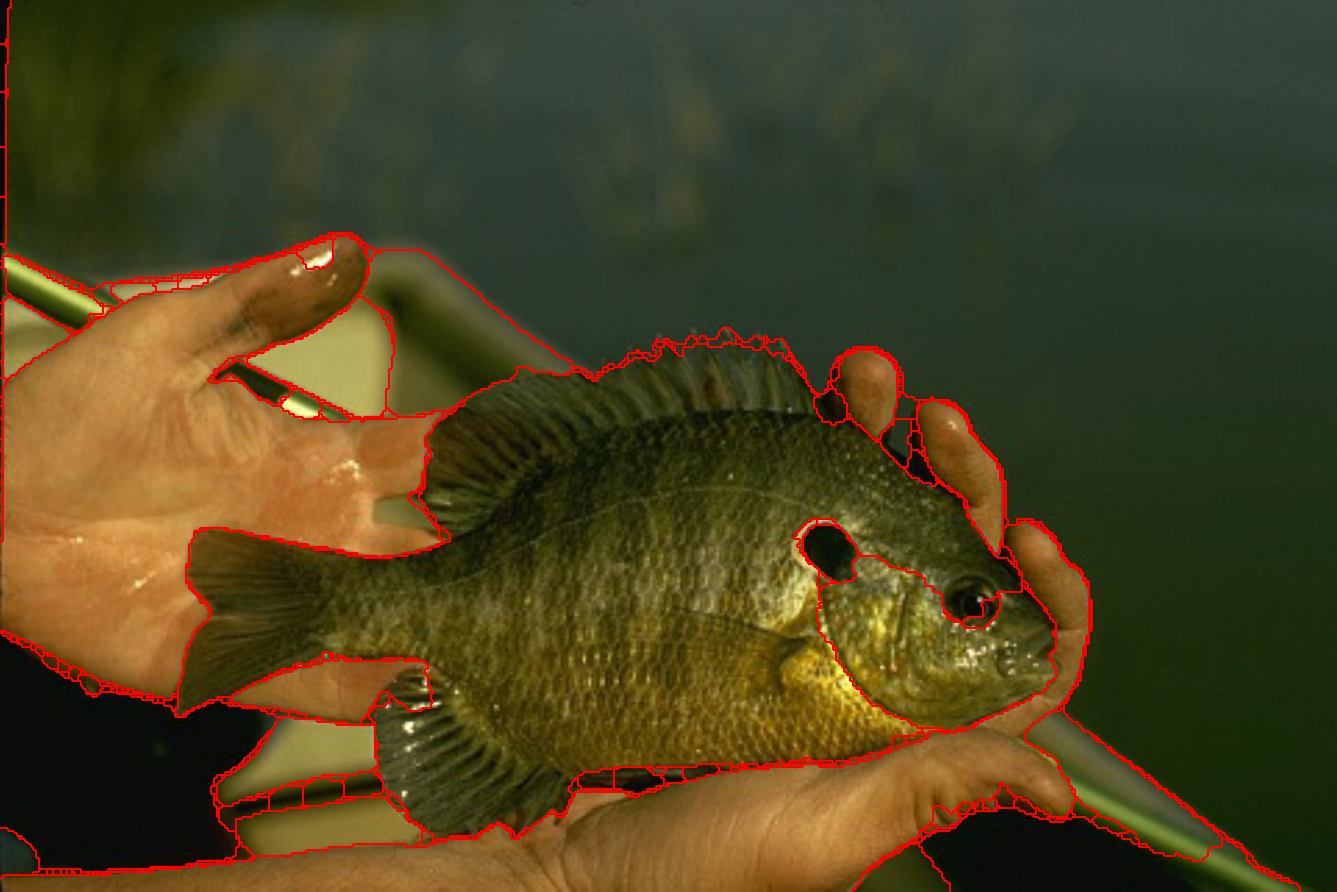}\\
\includegraphics[width=0.195\textwidth]{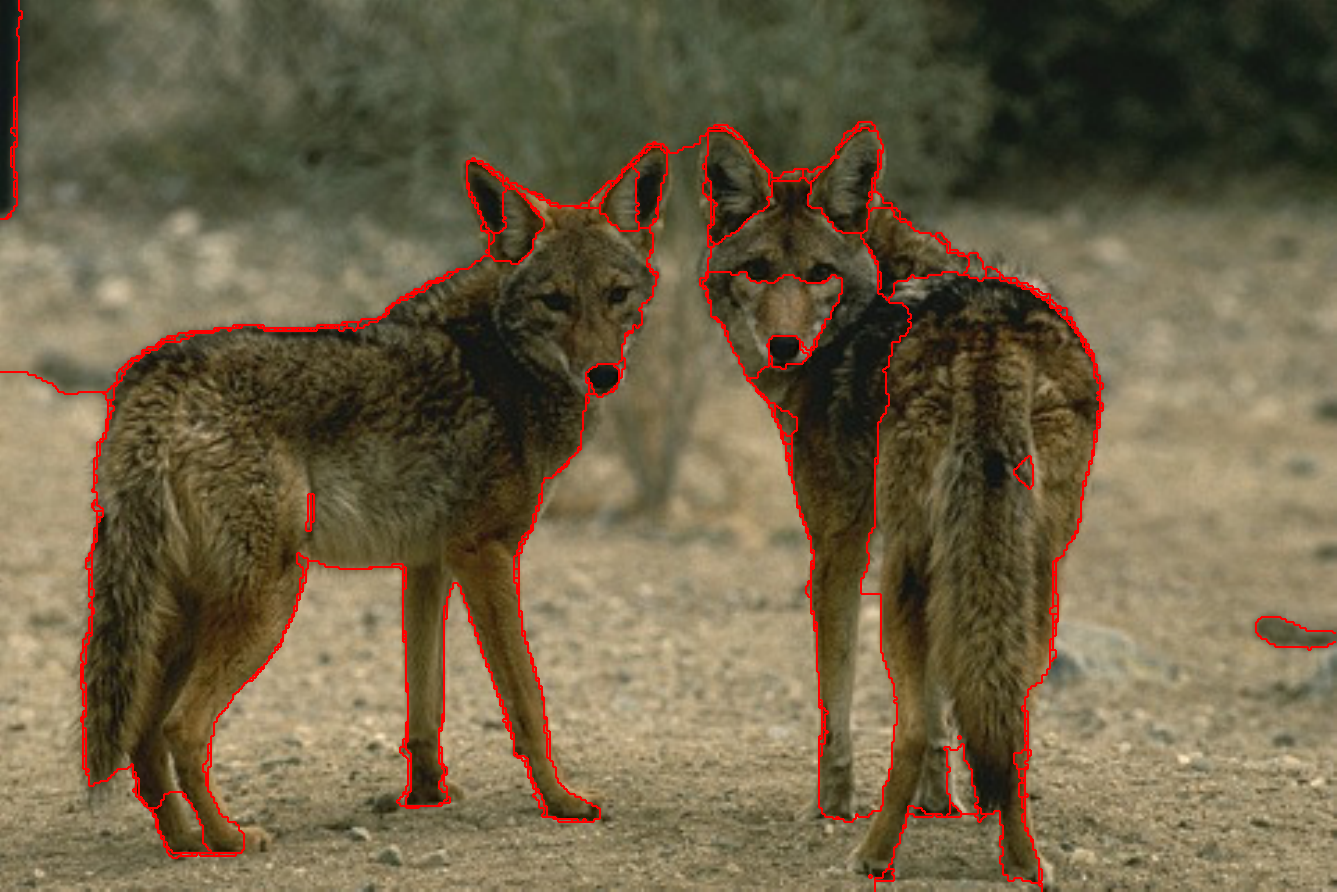}\hfill
\includegraphics[width=0.195\textwidth]{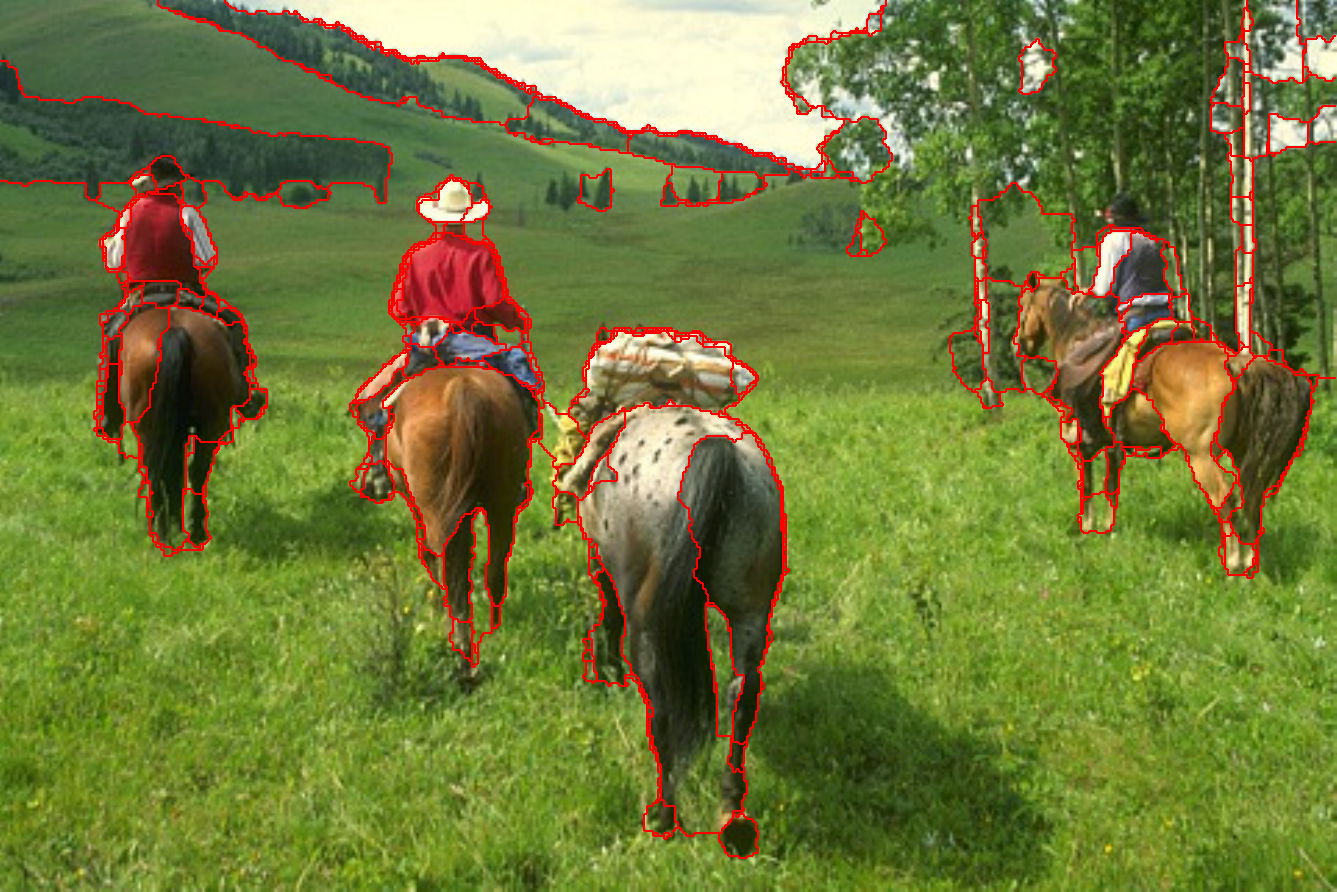}\hfill
\includegraphics[width=0.195\textwidth]{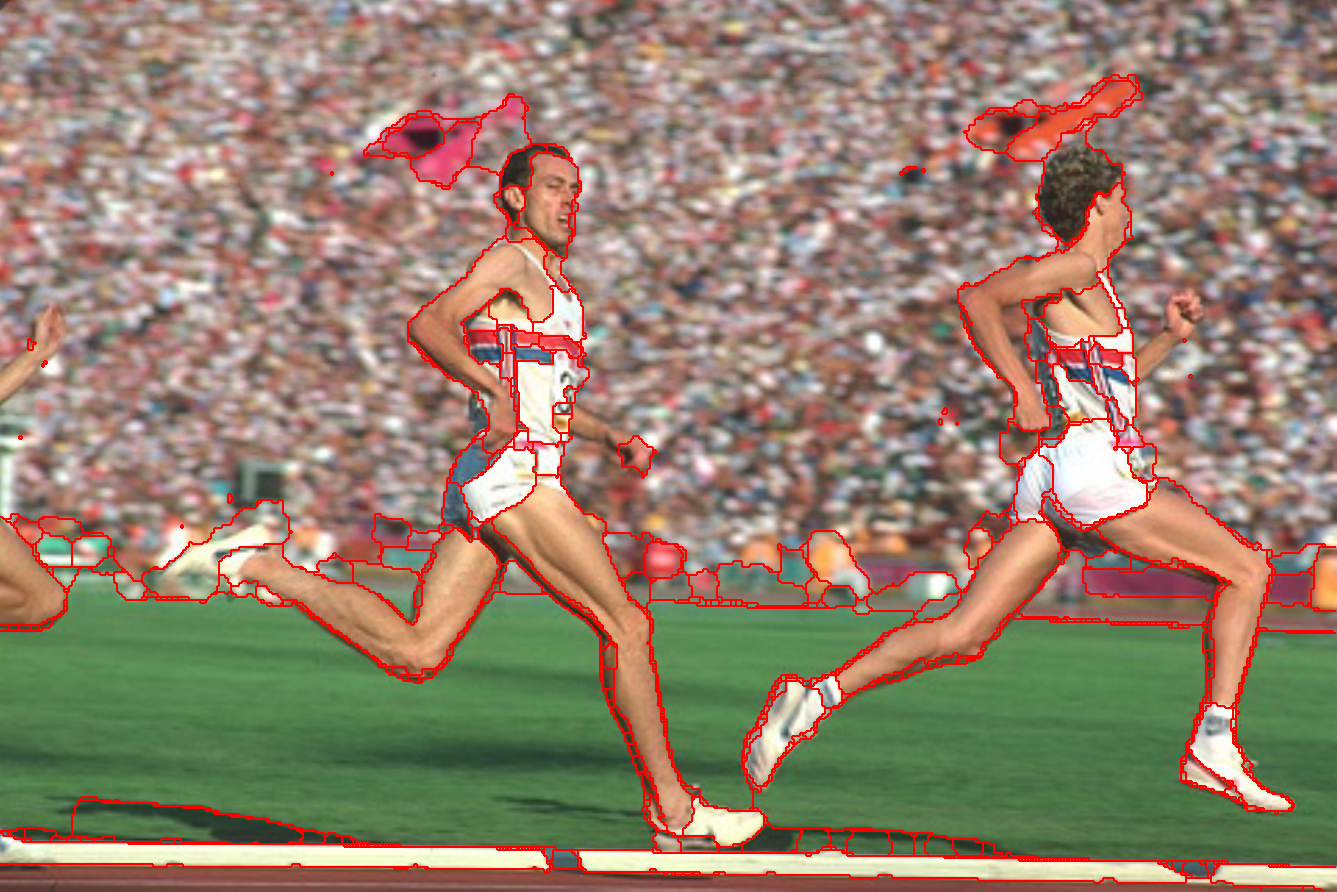}\hfill
\includegraphics[width=0.195\textwidth]{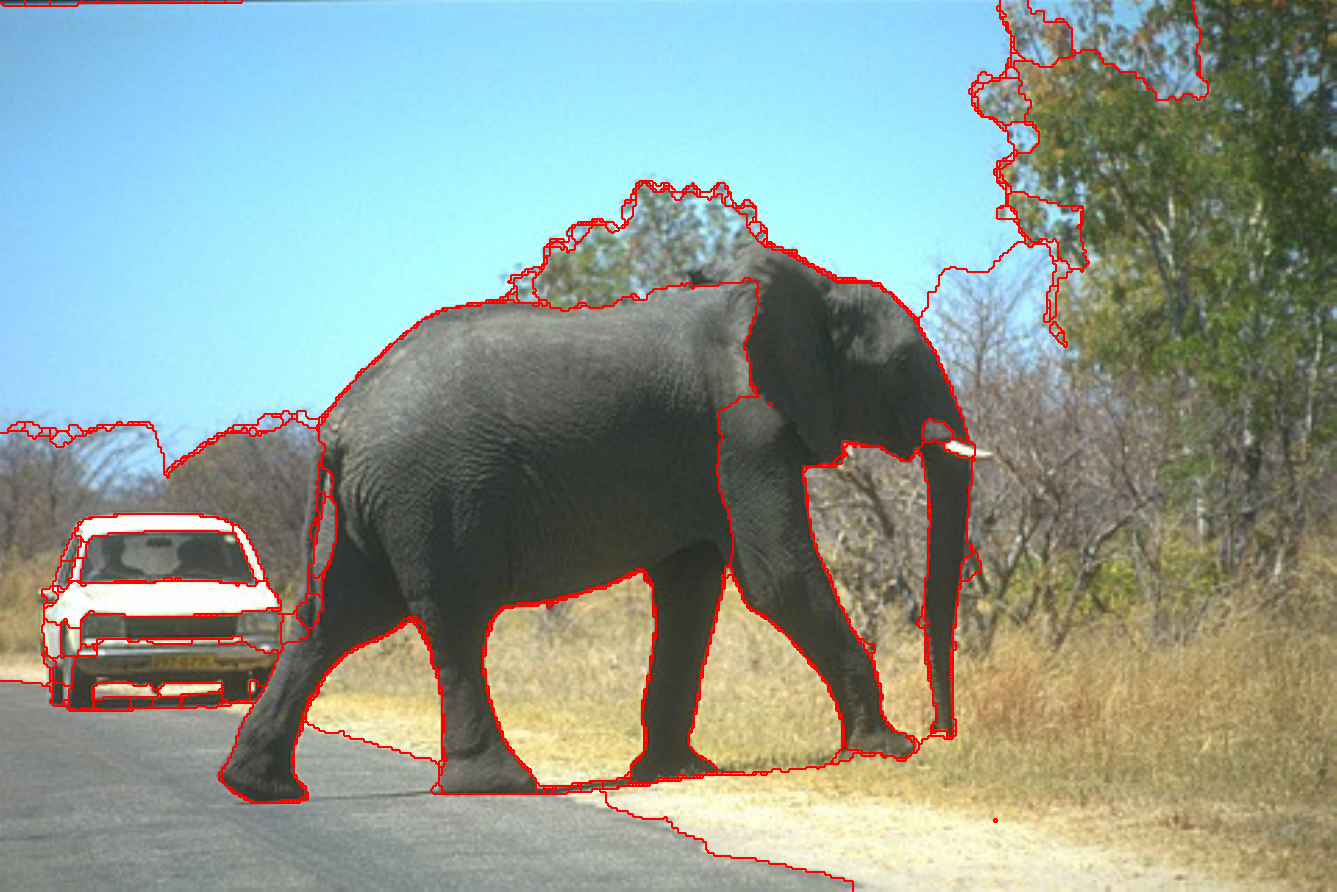}\hfill
\includegraphics[width=0.195\textwidth]{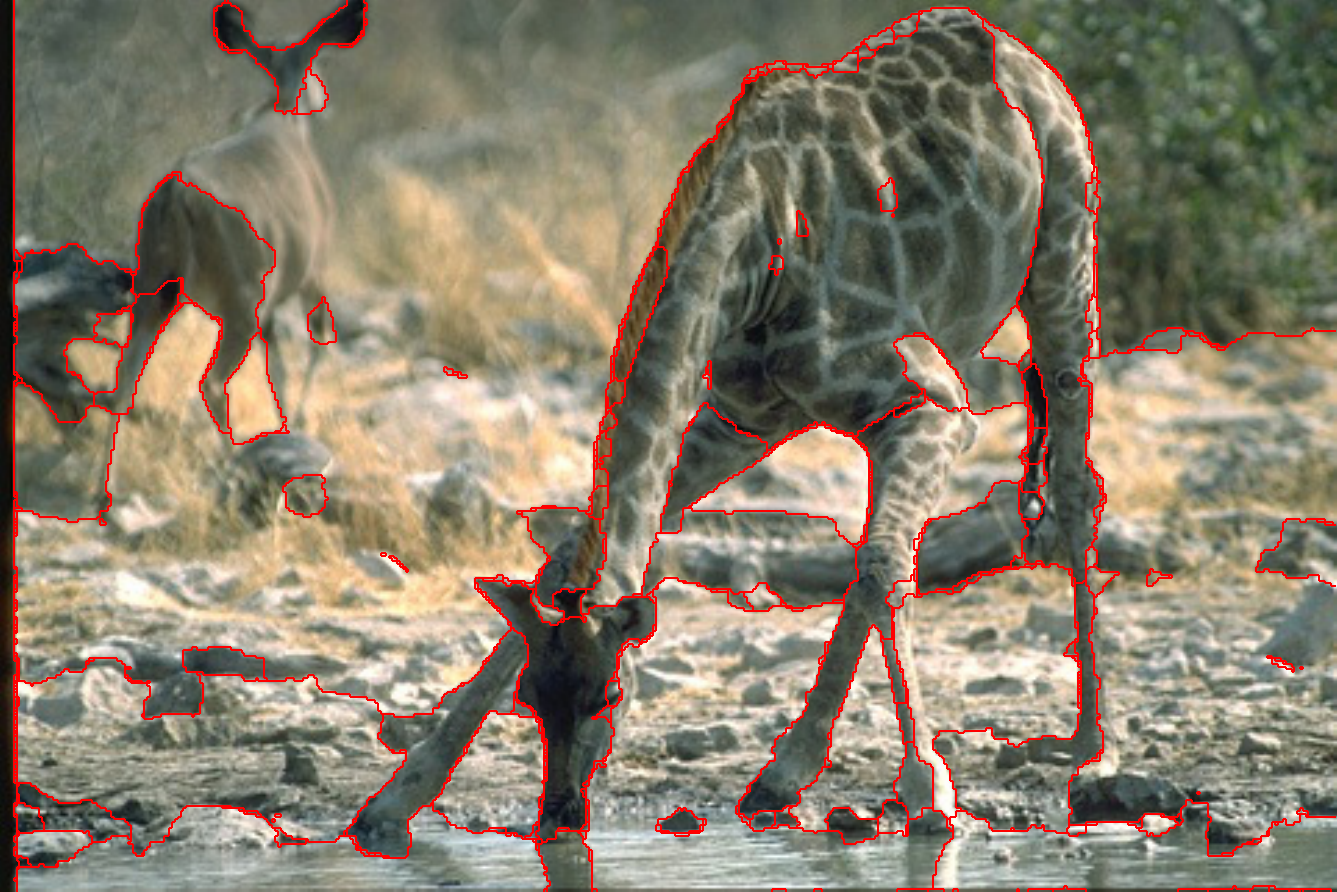}
\\[0.1ex]
\includegraphics[width=0.498\textwidth]{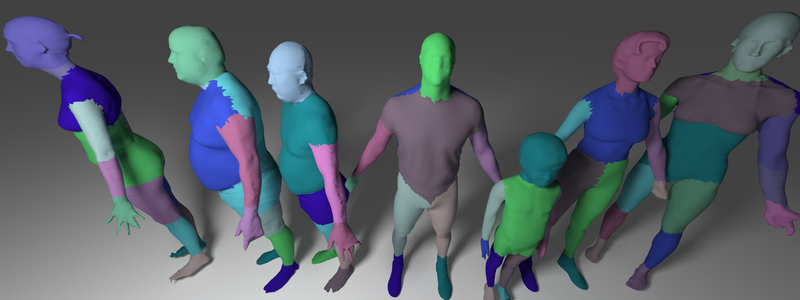}\hfill
\includegraphics[width=0.498\textwidth]{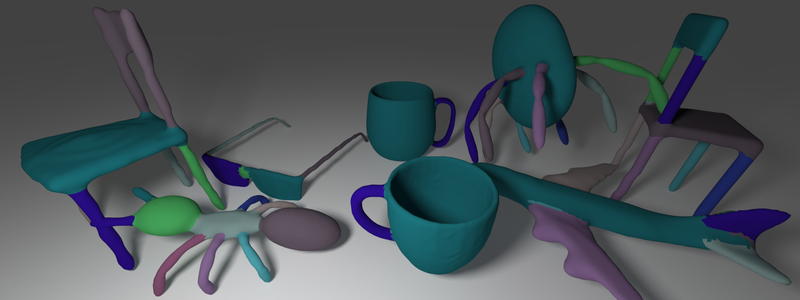}\\[0.1ex]
\includegraphics[width=0.498\textwidth]{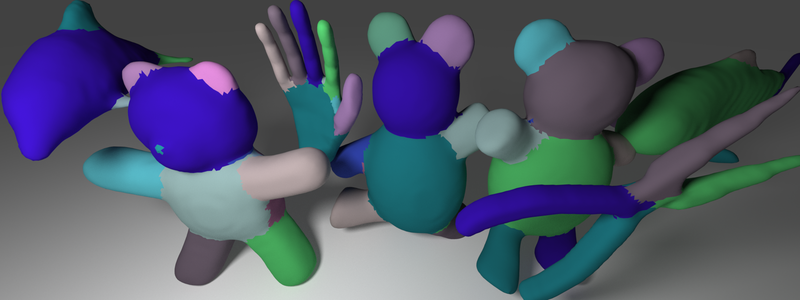}\hfill
\includegraphics[width=0.498\textwidth]{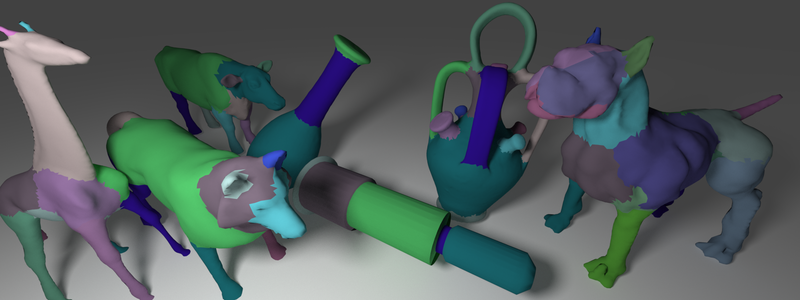}
\caption{Depicted above is a sample of decompositions of images and meshes found by solving an instance of the Lifted Multicut Problem (LMP) using Alg.~\ref{alg:KL_outer}.
Included are some cases where this approach fails.}
\label{fig:final}
\end{figure*}

We now apply our formulations and algorithms without any changes to the Mesh Segmentation Problem
\cite{Chen2009}.
For the LMP and Alg.~\ref{alg:KL_outer},
we obtain state-of-the-art results.

\textbf{Setup.}
The Princeton Segmentation Benchmark
\cite{Chen2009}
consists of 19 classes, ranging from humans to man-made objects, each containing 20 meshes. 
Manual segmentations of these meshes provide us with a ground truth for evaluation and supervised learning. 
We compute informative features known to provide good results in previous work~\cite{Kalogerakis2010,Benhabiles2011,Xie2014}: curvatures (minimum, maximum, Gaussian and mean) computed at two different scales, shape diameter~\cite{Shapira2008} and dihedral angle. 
Except the dihedral angle, which is computed for each edge, the curvatures and shape diameter are computed for each vertex of the mesh. 
To derive each of these criteria for an edge, we consider (1) its mean value over the two vertices of the edge, (2) its difference between the two vertices at each side of the edge and (3) the difference between its mean values computed on 1-ring neighborhoods at each side of the edge. This last combination provides additional robustness and multi-resolution behavior. 
This way, we obtain a 28-dimensional vector, which is more compact and efficient than the hundreds of features used in prior work~\cite{Kalogerakis2010,Xie2014}.
Probabilities of edges being cut are learned from the ground truth using a Random Forest classifier, through a leave-one-out experiment similarly to previous work~\cite{Kalogerakis2010,Xie2014}.
We apply Alg.~\ref{alg:KL_outer} on the dual graph of the mesh (one node per triangle), varying the prior probabilities $p^*$ of neighboring triangles being in distinct components, and $d^* \in \{60, 70, 80, 100\}$.

\textbf{Results.}
An evaluation in terms of Rand's index (RI) and the VI is shown in Tab.~\ref{tab:mesh}. 
The results are slightly better than~\cite{Zhanga,Kin-ChungAu2011} and close to those of~\cite{Kalogerakis2010}. 
However, Kalogerakis et al.~\cite{Kalogerakis2010} require a semantic labeling of the ground-truth, while our multicut formulation only requires boundary information. 
The median computation time per model is resp.~51 seconds for the lifting and 59 seconds for 
Alg.~\ref{alg:KL_outer} 
on an Intel i7 Pentium laptop computer operating at 2.20 GHz. 
Graphs have a median of 18000 nodes and 27000 edges.
A sample of our results is shown in Fig.~\ref{fig:final}. 
Varying the prior probability $p^*$ of cuts allows for controlling the amount of over or under-segmentation, as shown in Fig.~\ref{fig:meshbias}.

\section{Conclusion}
We have introduced a generalization of the Minimum Cost Multicut Problem (MP), 
the Minimum Cost Lifted Multicut Problem (LMP),
which overcomes limitations of the MP in applications to image and mesh segmentation.
We have defined and implemented two efficient algortihms (primal feasible heuristics) applicable to the MP and the LMP.
We have assessed both algorithms in conjunction with both optimization problems in applications 
to image decomposition (BSDS-500 benchmark \cite{arbelaez-2011})
and mesh decomposition (Princeton benchmark \cite{Chen2009}).
In both applications, we have found solutions that do not differ significantly from the state of the art.
This suggests that the LMP is a useful formulation of graph decomposition problems in vision.

\textbf{Acknowledgments.}
M.K.~and T.B.~acknowledge funding by the ERC Starting Grant Video\-Learn.
N.B.~and G.L.~acknowledge support by the Agence Nationale de la Recherche (ANR) through the CrABEx project (ANR-13-CORD-0013).

{\small\bibliographystyle{ieee}\bibliography{manuscript}}

\begin{thebibliography}{10}\itemsep=-1pt

\bibitem{our-code}
\url{http://www.andres.sc/graph.html}.

\bibitem{alush-2012}
A.~Alush and J.~Goldberger.
\newblock Ensemble segmentation using efficient integer linear programming.
\newblock {\em TPAMI}, 34(10):1966--1977, 2012.

\bibitem{andres-2015}
B.~Andres.
\newblock Lifting of multicuts.
\newblock {\em CoRR}, abs/1503.03791, 2015.

\bibitem{andres-2012-opengm}
B.~Andres, T.~Beier, and J.~H. Kappes.
\newblock {OpenGM}: {A} {C++} library for discrete graphical models.
\newblock {\em CoRR}, abs/1206.0111, 2012.

\bibitem{andres-2011}
B.~Andres, J.~H. Kappes, T.~Beier, U.~K\"othe, and F.~A. Hamprecht.
\newblock Probabilistic image segmentation with closedness constraints.
\newblock In {\em ICCV}, 2011.

\bibitem{andres-2012}
B.~Andres, T.~Kr{\"o}ger, K.~L. Briggman, W.~Denk, N.~Korogod, G.~Knott,
  U.~K{\"o}the, and F.~A. Hamprecht.
\newblock Globally optimal closed-surface segmentation for connectomics.
\newblock In {\em ECCV}, 2012.

\bibitem{andres-2013}
B.~Andres, J.~Yarkony, B.~S. Manjunath, S.~Kirchhoff, E.~Turetken, C.~Fowlkes,
  and H.~Pfister.
\newblock Segmenting planar superpixel adjacency graphs w.r.t.~non-planar
  superpixel affinity graphs.
\newblock In {\em EMMCVPR}, 2013.

\bibitem{arbelaez-2011}
P.~Arbel\'{a}ez, M.~Maire, C.~Fowlkes, and J.~Malik.
\newblock Contour detection and hierarchical image segmentation.
\newblock {\em TPAMI}, 33(5):898--916, 2011.

\bibitem{APBMM2014}
P.~Arbel\'{a}ez, J.~Pont-Tuset, J.~Barron, F.~Marques, and J.~Malik.
\newblock Multiscale combinatorial grouping.
\newblock In {\em CVPR}, 2014.

\bibitem{bagon-2011}
S.~Bagon and M.~Galun.
\newblock Large scale correlation clustering optimization.
\newblock {\em CoRR}, abs/1112.2903, 2011.

\bibitem{bansal-2004}
N.~Bansal, A.~Blum, and S.~Chawla.
\newblock Correlation clustering.
\newblock {\em Machine Learning}, 56(1--3):89--113, 2004.

\bibitem{beier-2015}
T.~Beier, F.~A. Hamprecht, and J.~H. Kappes.
\newblock Fusion moves for correlation clustering.
\newblock In {\em CVPR}, 2015.

\bibitem{beier-2014}
T.~Beier, T.~Kr{\"o}ger, J.~H. Kappes, U.~K{\"o}the, and F.~A. Hamprecht.
\newblock {Cut, Glue \& Cut}: A fast, approximate solver for multicut
  partitioning.
\newblock In {\em CVPR}, 2014.

\bibitem{Benhabiles2011}
H.~Benhabiles, G.~Lavou\'{e}, J.-P. Vandeborre, and M.~Daoudi.
\newblock {Learning Boundary Edges for 3D-Mesh Segmentation}.
\newblock {\em Computer Graphics Forum}, 30(8):2170--2182, 2011.

\bibitem{bertrasius-2015}
G.~Bertasius, J.~Shi, and L.~Torresani.
\newblock Deepedge: A multi-scale bifurcated deep network for top-down contour
  detection.
\newblock In {\em CVPR}, 2015.

\bibitem{Chen2009}
X.~Chen, A.~Golovinskiy, and T.~Funkhouser.
\newblock {A benchmark for 3D mesh segmentation}.
\newblock {\em TOG}, 28(3):73, 2009.

\bibitem{chopra-1993}
S.~Chopra and M.~Rao.
\newblock The partition problem.
\newblock {\em Mathematical Programming}, 59(1--3):87--115, 1993.

\bibitem{demaine-2006}
E.~D. Demaine, D.~Emanuel, A.~Fiat, and N.~Immorlica.
\newblock Correlation clustering in general weighted graphs.
\newblock {\em Theoretical Computer Science}, 361(2--3):172--187, 2006.

\bibitem{deza-1997}
M.~M. Deza and M.~Laurent.
\newblock {\em Geometry of Cuts and Metrics}.
\newblock Springer, 1997.

\bibitem{DollarICCV13edges}
P.~Doll\'ar and C.~L. Zitnick.
\newblock Structured forests for fast edge detection.
\newblock In {\em ICCV}, 2013.

\bibitem{fowlkes-2015}
S.~Hallman and C.~Fowlkes.
\newblock Oriented edge forests for boundary detection.
\newblock In {\em CVPR}, 2015.

\bibitem{hopcroft-1973}
J.~Hopcroft and R.~Tarjan.
\newblock Algorithm 447: Efficient algorithms for graph manipulation.
\newblock {\em Commun. ACM}, 16(6):372--378, 1973.

\bibitem{crisp_boundaries}
P.~Isola, D.~Zoran, D.~Krishnan, and E.~H. Adelson.
\newblock Crisp boundary detection using pointwise mutual information.
\newblock In {\em ECCV}, 2014.

\bibitem{Kalogerakis2010}
E.~Kalogerakis and A.~Hertzmann.
\newblock {Learning 3D mesh segmentation and labeling}.
\newblock {\em TOG}, 29(4):102, 2010.

\bibitem{kappes-2015}
J.~H. Kappes, B.~Andres, F.~A. Hamprecht, C.~Schn\"orr, S.~Nowozin, D.~Batra,
  S.~Kim, B.~X. Kausler, T.~Kr\"oger, J.~Lellmann, N.~Komodakis,
  B.~Savchynskyy, and C.~Rother.
\newblock A comparative study of modern inference techniques for structured
  discrete energy minimization problems.
\newblock {\em IJCV}, 2015.

\bibitem{kappes-2011}
J.~H. Kappes, M.~Speth, B.~Andres, G.~Reinelt, and C.~Schn\"orr.
\newblock Globally optimal image partitioning by multicuts.
\newblock In {\em EMMCVPR}, 2011.

\bibitem{kappes-2013-arxiv}
J.~H. Kappes, M.~Speth, G.~Reinelt, and C.~Schn{\"{o}}rr.
\newblock Higher-order segmentation via multicuts.
\newblock {\em CoRR}, abs/1305.6387, 2013.

\bibitem{kappes-2015-ssvm}
J.~H. Kappes, P.~Swoboda, B.~Savchynskyy, T.~Hazan, and C.~Schn\"orr.
\newblock Probabilistic correlation clustering and image partitioning using
  perturbed multicuts.
\newblock In {\em SSVM}, 2015.

\bibitem{kernighan-1970}
B.~W. Kernighan and S.~Lin.
\newblock An efficient heuristic procedure for partitioning graphs.
\newblock {\em Bell Systems Technical Journal}, 49:291--307, 1970.

\bibitem{keuper-2015b}
M.~Keuper, B.~Andres, and T.~Brox.
\newblock Motion trajectory segmentation via minimum cost multicuts.
\newblock In {\em ICCV}, 2015.

\bibitem{kim-2011}
S.~Kim, S.~Nowozin, P.~Kohli, and C.~Yoo.
\newblock Higher-order correlation clustering for image segmentation.
\newblock In {\em NIPS}, 2011.

\bibitem{kim-2014}
S.~Kim, C.~Yoo, S.~Nowozin, and P.~Kohli.
\newblock Image segmentation using higher-order correlation clustering.
\newblock {\em TPAMI}, 36:1761--1774, 2014.

\bibitem{Kin-ChungAu2011}
O.~{Kin-Chung Au}, Y.~Zheng, M.~Chen, P.~Xu, and C.-L. Tai.
\newblock {Mesh Segmentation with Concavity-Aware Fields.}
\newblock {\em IEEE Transactions on Visualization and Computer Graphics},
  18(7):1125 -- 1134, 2011.

\bibitem{meila-2007}
M.~Meil\u{a}.
\newblock Comparing clusterings---an information based distance.
\newblock {\em J. Multivar. Anal.}, 98(5):873--895, 2007.

\bibitem{nowozin-2009}
S.~Nowozin and S.~Jegelka.
\newblock Solution stability in linear programming relaxations: Graph
  partitioning and unsupervised learning.
\newblock In {\em ICML}, 2009.

\bibitem{Shapira2008}
L.~Shapira, A.~Shamir, and D.~Cohen-Or.
\newblock {Consistent mesh partitioning and skeletonisation using the shape
  diameter function}.
\newblock {\em The Visual Computer}, 24(4):249--259, 2008.

\bibitem{shi-2000}
J.~Shi and J.~Malik.
\newblock Normalized cuts and image segmentation.
\newblock {\em TPAMI}, 22(8):888--905, 2000.

\bibitem{Theologou2015}
P.~Theologou, I.~Pratikakis, and T.~Theoharis.
\newblock {A comprehensive overview of methodologies and performance evaluation
  frameworks in 3D mesh segmentation}.
\newblock {\em CVIU}, 135:49--82, 2015.

\bibitem{Xie2014}
Z.~Xie, K.~Xu, L.~Liu, and Y.~Xiong.
\newblock {3D Shape Segmentation and Labeling via Extreme Learning Machine}.
\newblock {\em Computer Graphics Forum}, 33(5), 2014.

\bibitem{yarkony-2012}
J.~Yarkony, A.~Ihler, and C.~Fowlkes.
\newblock Fast planar correlation clustering for image segmentation.
\newblock In {\em ECCV}, 2012.

\bibitem{yarkony-2015}
J.~Yarkony, C.~Zhang, and C.~Fowlkes.
\newblock Hierarchical planar correlation clustering for cell segmentation.
\newblock In {\em EMMCVPR}, 2015.

\bibitem{Zhanga}
J.~Zhang, J.~Zheng, C.~Wu, and {Jianfei Cai}.
\newblock {Variational Mesh Decomposition}.
\newblock {\em TOG}, 31(3), 2012.

\end{thebibliography}

\end{document}